\documentclass{article}

\PassOptionsToPackage{sort&compress,numbers}{natbib}
\usepackage[preprint]{neurips_2025}
\usepackage{natbib}

\usepackage[utf8]{inputenc} %
\usepackage[T1]{fontenc}    %
\usepackage{url}            %
\usepackage{booktabs}       %
\usepackage{amsfonts}       %
\usepackage{nicefrac}       %
\usepackage{microtype}      %
\usepackage[percent]{overpic}
\usepackage{pict2e}

\usepackage{amsmath}
\usepackage{amssymb}
\usepackage{mathtools}
\usepackage{amsthm}
\usepackage{stmaryrd}
\usepackage{enumitem}

\usepackage{tabularx}
\usepackage{makecell}

\usepackage{url}            %
\usepackage{booktabs}       %
\usepackage{amsfonts}       %
\usepackage{nicefrac}       %
\usepackage{microtype}      %
\usepackage{multirow}       %
\usepackage{algorithm}
\usepackage{algorithmic}

\usepackage{wrapfig}
\usepackage{comment}
\usepackage{amsmath,amssymb} %
\usepackage{symbols}
\usepackage{multirow}
\usepackage{pifont}%

\usepackage{amsmath,amsfonts,bm}

\def\1{\bm{1}}

\def\sp{space}

\def\rvc{{\mathbf{c}}}

\def\rvx{{\mathbf{x}}}
\def\rvy{{\mathbf{y}}}
\def\rvz{{\mathbf{z}}}

\def\rmA{{\mathbf{A}}}

\def\rmK{{\mathbf{K}}}

\def\rmQ{{\mathbf{Q}}}

\def\rmV{{\mathbf{V}}}
\def\rmW{{\mathbf{W}}}

\def\vc{{\bm{c}}}

\def\vn{{\bm{n}}}

\def\vx{{\bm{x}}}
\def\vy{{\bm{y}}}
\def\vz{{\bm{z}}}

\def\mE{{\bm{E}}}

\def\mI{{\bm{I}}}

\DeclareMathAlphabet{\mathsfit}{\encodingdefault}{\sfdefault}{m}{sl}
\SetMathAlphabet{\mathsfit}{bold}{\encodingdefault}{\sfdefault}{bx}{n}
\newcommand{\tens}[1]{\bm{\mathsfit{#1}}}

\def\tD{{\tens{D}}}
\def\tE{{\tens{E}}}

\def\gD{{\mathcal{D}}}
\def\gE{{\mathcal{E}}}

\def\gL{{\mathcal{L}}}

\def\gN{{\mathcal{N}}}

\def\sR{{\mathbb{R}}}

\usepackage{color}
\usepackage{soul}
\usepackage{multirow}
\usepackage{xcolor}

\definecolor{darkred}{rgb}{0.7,0.1,0.1}
\definecolor{darkgreen}{rgb}{0.1,0.7,0.1}
\definecolor{cyan}{rgb}{0.7,0.0,0.7}
\definecolor{dblue}{rgb}{0.2,0.2,0.8}
\definecolor{maroon}{rgb}{0.76,.13,.28}
\definecolor{burntorange}{rgb}{0.81,.33,0}
\definecolor{tealblue}{rgb}{0.212,0.459, 0.533}
\definecolor{myyellow}{rgb}{0.8627451 , 0.75294118, 0.20784314]}

\definecolor{mypink}{rgb}{0.93359375, 0.62109375, 0.83984375}

\definecolor{pp}{rgb}{0.43921569, 0.18823529, 0.62745098}
\definecolor{rr}{rgb}{0.5254902 , 0.00784314, 0.12941176}
\definecolor{bb}{rgb}{0.09019608, 0.23529412, 0.37647059}
\definecolor{yy}{rgb}{0.49803922, 0.3372549 , 0.0}
\definecolor{gg}{rgb}{0.02352941, 0.3372549 , 0.17647059}

\definecolor{lightred}{rgb}{0.9,0.4,0.4}

\newlength\savewidth

\definecolor{turquoise}{cmyk}{0.65,0,0.1,0.1}
\definecolor{purple}{rgb}{0.65,0,0.65}
\definecolor{darkgreen}{rgb}{0.0, 0.5, 0.0}
\definecolor{darkred}{rgb}{0.5, 0.0, 0.0}
\definecolor{darkblue}{rgb}{0.0, 0.0, 0.5}
\definecolor{blue}{rgb}{0.0, 0.0, 1.0}
\definecolor{orange}{rgb}{1.0,0.5,0.0}

\usepackage[multiple]{footmisc}
\usepackage{wrapfig}

\definecolor{mybrown}{rgb}{0.87058824, 0.56078431, 0.01960784}
\definecolor{myblue}{rgb}{0.3372549 , 0.70588235, 0.91372549}
\definecolor{mypurple}{rgb}{0.8, 0.47058824, 0.7372549 }
\definecolor{myorange}{rgb}{0.835, 0.368, 0}
\definecolor{mygreen}{rgb}{0.00784314, 0.61960784, 0.45098039}
\definecolor{mygt}{rgb}{0.0078125 , 0.57421875, 0.40625}
\definecolor{mysp}{rgb}{0.84765625, 0.515625  , 0.0234375}
\definecolor{mycitecolor}{rgb}{0,0.08,0.45}
\definecolor{mygr}{rgb}{0.9607,0.9607,0.9607}
\definecolor{myoo}{rgb}{0.992,0.9176,0.9019}

\usepackage{listings}

\definecolor{codegray}{gray}{0.95}
\definecolor{keywordblue}{rgb}{0.13,0.29,0.53}
\definecolor{stringred}{rgb}{0.58,0,0.22}
\definecolor{commentgreen}{rgb}{0,0.5,0}

\definecolor{myrr}{HTML}{AE031A}
\definecolor{mybb}{HTML}{0155B3}

\usepackage{pifont}
\usepackage[T1]{fontenc}
\usepackage[utf8]{inputenc}
\usepackage{pmboxdraw}
\usepackage{thm-restate}
\definecolor{iccvblue}{rgb}{0.21,0.49,0.74}
\usepackage[pagebackref,breaklinks,colorlinks,allcolors=iccvblue]{hyperref}

\usepackage{url}            %
\usepackage{booktabs}       %
\usepackage{amsfonts}       %
\usepackage{nicefrac}       %
\usepackage{microtype}      %
\usepackage{multirow}       %
\usepackage{algorithm}
\usepackage{algorithmic}
\usepackage{mathtools}%

\usepackage{arydshln}

\usepackage{symbols}

\newcommand{\myparagraph}[1]{\vspace*{0pt}{\bf\noindent #1}}
\newcommand{\ours}{\textbf{DarkDiff}\xspace}

\title{
DarkDiff: Advancing Low-Light Raw Enhancement by \\Retasking Diffusion Models for Camera ISP
}
\author{
Amber Yijia Zheng$^{1}$\thanks{Work was done during an internship at Apple Inc.} \quad Yu Zhang$^{2}$ \quad Jun Hu$^{2}$ \quad Raymond A. Yeh$^{1}$\thanks{Equal advising.} \quad Chen Chen$^{2}$\footnotemark[2] \\
\textsuperscript{1}Department of Computer Science, Purdue University \hspace{10mm} \textsuperscript{2}Apple Inc. 
}

\begin{document}
\maketitle

\begin{figure*}[h]
\small
\vspace{-.5cm}
\setlength{\tabcolsep}{0pt}
\renewcommand{\arraystretch}{0.55}
\centering
    \begin{tabular}{c@{\hspace{1mm}}c@{\hspace{1mm}}c@{\hspace{1mm}}c}
            Noisy Input & Reference  & Exposure Diffusion~\cite{wang2023exposurediffusion} & \bf Ours \\
        \includegraphics[width=0.22\textwidth]{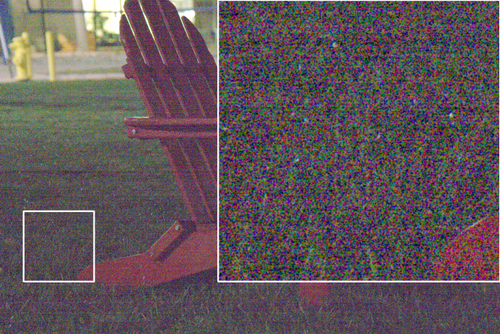}& 
        \includegraphics[width=0.22\textwidth]{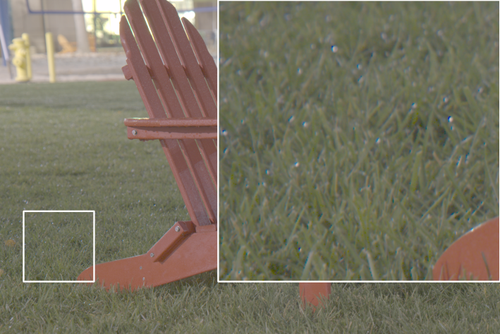} & 
        \includegraphics[width=0.22\textwidth]{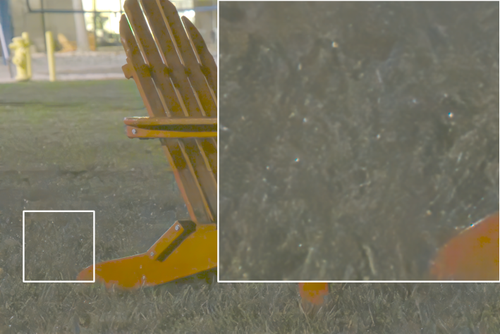} & 
        \includegraphics[width=0.22\textwidth]{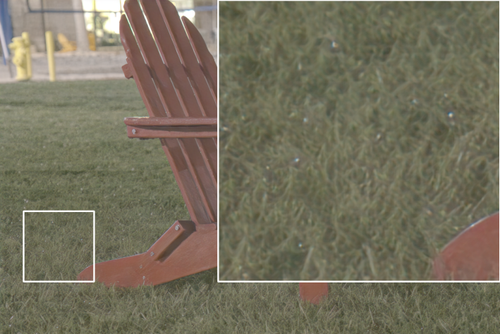} \\
        \includegraphics[width=0.22\textwidth]{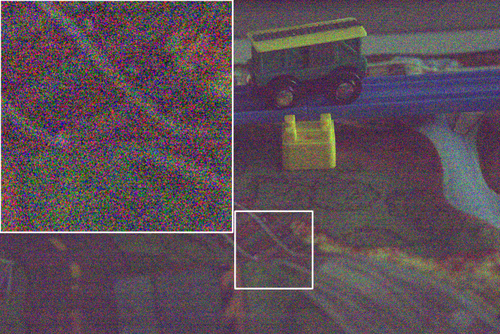}& 
        \includegraphics[width=0.22\textwidth]{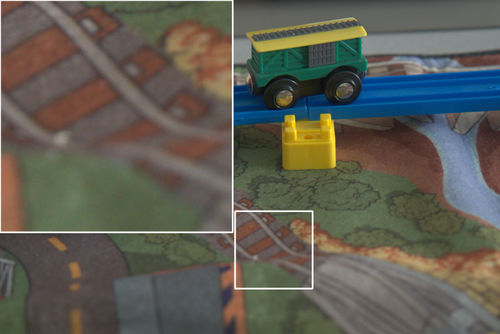} & 
        \includegraphics[width=0.22\textwidth]{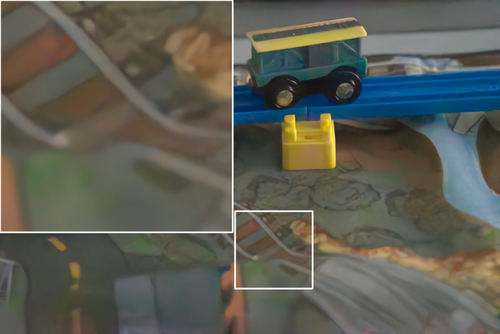} & 
        \includegraphics[width=0.22\textwidth]{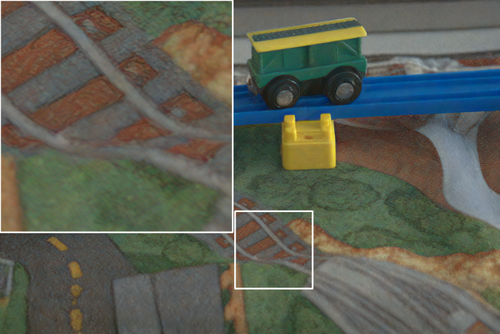}
    \end{tabular}
    \vspace{-0.22cm}
    \caption{Comparisons of low-light raw image enhancement results. The two input raw images were captured at night with only 0.1s and 0.033s exposure time by a Sony A7SII camera~\cite{chen2018learning}. 
    A digital gain of 300 and gamma correction have been applied for visualization. With sharp and vivid content, our results are comparable to the reference images captured with 300 times longer exposure on a tripod. 
    }
    \label{fig:teaser}
\end{figure*}

\begin{abstract}

High-quality photography in extreme low-light conditions is challenging but impactful for digital cameras. With advanced computing hardware, traditional camera image signal processor (ISP) algorithms are gradually being replaced by efficient deep networks that enhance noisy raw images more intelligently. However, existing regression-based models often minimize pixel errors and result in oversmoothing of low-light photos or deep shadows. Recent work has attempted to address this limitation by training a diffusion model from scratch, yet those models still struggle to recover sharp image details and accurate colors. We introduce a novel framework to enhance low-light raw images by retasking pre-trained generative diffusion models with the camera ISP. Extensive experiments demonstrate that our method outperforms the state-of-the-art in perceptual quality across three challenging low-light raw image benchmarks.

\end{abstract}

\section{Introduction}
\label{sec:intro}
Enhancing noisy \textit{raw} images to vivid, clean sRGB images is typically managed by camera image signal processor (ISP) algorithms in digital cameras, %
which include image denoising, demosaicing, white balance, color correction, gamma correction, and tone mapping, \etc, have become increasingly sophisticated over decades of development~\cite{hasinoff2016burst,liba2019handheld}. However, capturing images in extremely low-light conditions, where sensor photon count is very low, remains challenging for existing ISP algorithms. When these algorithms fail to recover details from highly noisy areas, oversmoothing or oil-painting appearances are common artifacts.

To improve traditional ISP algorithms, existing works mainly focused on regression-based deep networks to enhance the raw images~\cite{chen2018learning,wang2023exposurediffusion,jiniccv23led}. %
Seminal work SID~\cite{chen2018learning} collected a dataset of paired noisy raw and clean sRGB images to learn the camera ISP using deep networks, while other works~\cite{wei2020physics,zhang2023towards,abdelhamed2019noise,brooks2019unprocessing} studied different synthetic noise models to improve noise reduction in ISP.

Despite the progress, regression-based methods often encounter the problem of "regression toward the mean," which results in oversmoothed images. To mitigate this, recent work, such as ExposureDiffusion~\cite{wang2023exposurediffusion}, has shifted toward a generative approach to generate {\it plausible} content in areas with a large amount of noise. %
However, when image capturing conditions are extreme, ExposureDiffusion~\cite{wang2023exposurediffusion} still struggles to recover high-resolution image details and accurate colors; see~\figref{fig:teaser} for examples.

\begin{figure*}[t]
    \centering
    \includegraphics[width=\linewidth]{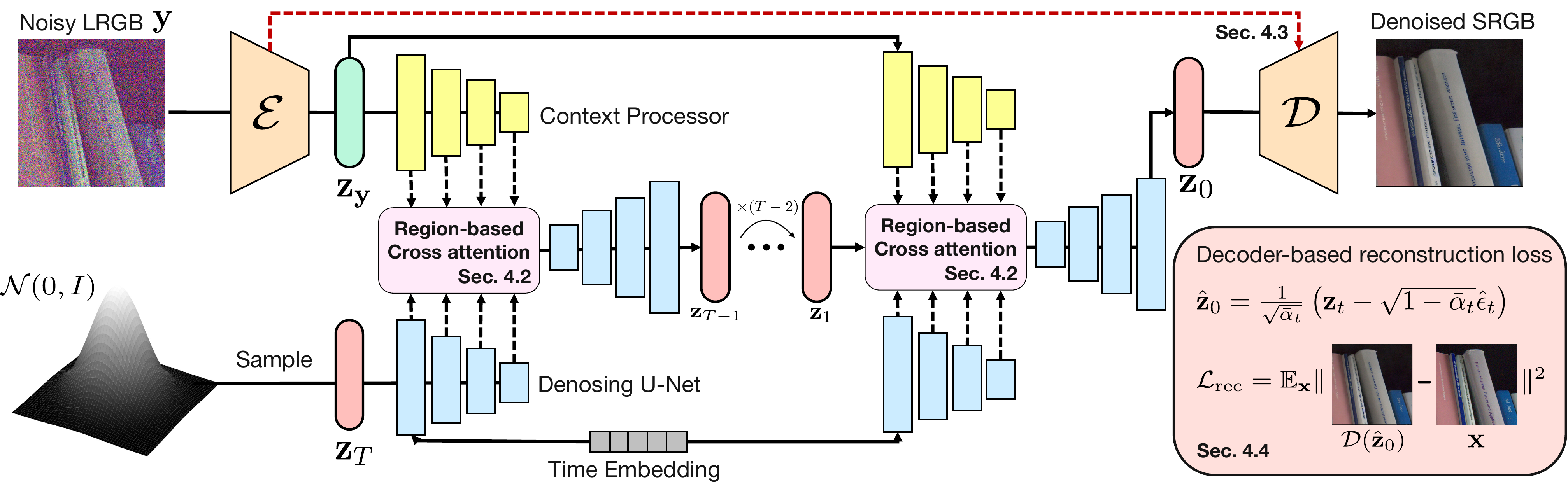}
    \vspace{-0.35cm}
    \caption{\textbf{Overview of the proposed DarkDiff pipeline.} 
    The noisy linear RBG (LRGB) image $\rvy$ is processed by the encoder $\gE$ to generate latent representations $\rvz_\rvy$. These representations and the Gaussian noise are fed into the Denoising U-Net, which integrates a region-based cross-attention between the noisy image and context from the pre-trained model to refine the latent variables $\rvz_t$. The decoder $\gD$ reconstructs the final clean SRGB image $\rvx$. The region-based cross-attention mechanism (\secref{sec:x-atten}) allows the model to leverage contextual information at each denoising step for better detail preservation and noise reduction.}
    \label{fig:pipeline}
    \vspace{-0.2cm}
\end{figure*}

In this work, we leverage the pre-trained StableDiffuion's~\cite{rombach2022high} generative capability to improve extremely low-light raw image enhancement. A new framework is carefully designed to overcome the limitations of existing deep network-based ISPs. The relatively mature functions in ISP, such as Bayer processing and white balance, are applied before the network. The proposed method learns to map noisy linear RGB images to vivid and clean sRGB outputs. An overview of the method is illustrated in~\figref{fig:pipeline}.

We conduct experiments on three challenging datasets, including SID~\cite{chen2018learning}, ELD~\cite{wei2020physics}, and LRD~\cite{zhang2023towards} for extreme low-light raw image enhancement. Results show that our approach outperforms the state-of-the-art (SOTA) in terms of overall perceptual quality (LPIPS~\cite{zhang2018unreasonable}) while achieving competitive performance in reconstruction fidelity (measured in PSNR). Additionally, qualitative results show that our approach produces visually more appealing images while preserving high fidelity.

{\bf\noindent Our contributions are as follows:}%
\begin{itemize}[topsep=0pt, leftmargin=16pt]
    \setlength{\itemsep}{0.0pt}
    \setlength{\parskip}{2.5pt}
    \item We introduce DarkDiff, a novel approach to retask a pre-trained diffusion model into the camera ISP. Our approach adapts the generative capability trained from large-scale internet images into low-light raw image enhancement, where no large-scale raw datasets are available. %
    \item With a novel architecture design for low-light raw image enhancement, DarkDiff incorporates three key components to effectively address over-smoothing, low fidelity, and color shifts, which are common in prior methods. 
    \item Comprehensive experiments demonstrate that DarkDiff achieves SOTA performance, outperforming baselines in the perceptual metric LPIPS and producing visually higher quality images. %
\end{itemize}

\section{Related Work}
\myparagraph{Low-light image enhancement.} One line of research studies the direct enhancement of sRGB images~\cite{guo2016lime,guo2023low,lore2017llnet,wang2023low,xu2023low,yi2023diff},~\ie, the output of the camera's Image Signal Processor (ISP). These works aim to map a low-light dark sRGB image to a properly exposed image under the assumption that the input images' noise was already addressed by the camera ISP. However, when the ISP does not fully remove the noise, the noise is easily amplified during the enhancement process.

To mitigate this noise implication, another line of work~\cite{chen2018learning,wei2020physics,zhang2023towards,wang2023exposurediffusion} enhances the image at its source, before being fully processed by ISP,~\ie, the camera raw data. These methods train end-to-end networks for raw-to-raw or raw-to-sRGB mapping. For instance, SID~\cite{chen2018learning} learns a U-Net with a regression loss on paired noisy raw images and clean sRGB images. ELD~\cite{wei2020physics} and LRD~\cite{zhang2023towards} introduce noise models to reduce the domain gap between synthetic and real images, while DNF~\cite{jincvpr23dnf} proposes a two-stage approach that sequentially performs denoising and color correction. However, these methods struggle in regions with extremely low SNR, where input data lacks sufficient information.

Recent works consider generative models to further improve image quality,~\eg, using diffusion models~\cite{wang2023exposurediffusion, jiang2023low,Jiang_2024_ECCV,nguyen2024diffusion,yi2023diff,hou2024global,he2023reti}. At a high level, these methods train diffusion models from scratch using small-scale paired low-light image datasets, which have very limited content generation capability under severe noise conditions. Closely related to our work, LDM-ISP~\cite{wen2023ldm} introduces DWT preprocessing and taming modules for low-light enhancement. Differently, we carefully designed the data pipeline to reduce the domain gap and introduced new conditioning modules to better utilize a pre-trained diffusion model.

\begin{figure*}[t]
    \centering
    \begin{minipage}[t]{0.48\linewidth}
        \centering
        \small
        \begin{tabular}{@{}c@{\hspace{0.5mm}}c@{\hspace{0.5mm}}c}
            Noisy Input & Reference  & LDM Output \\
            \begin{overpic}[width=0.33\linewidth]{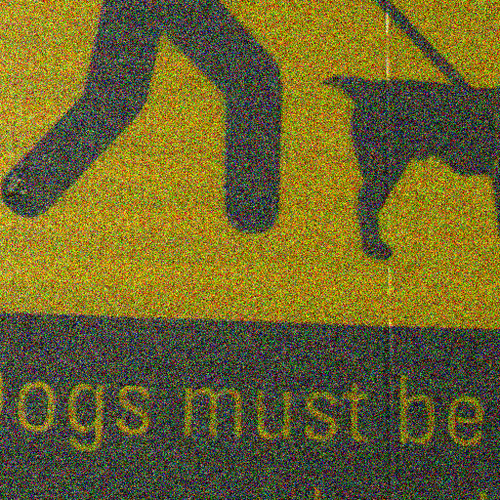}
            \put(78,70){\color{red}\oval(40,50)}
            \end{overpic} &
            \begin{overpic}[width=0.33\linewidth]{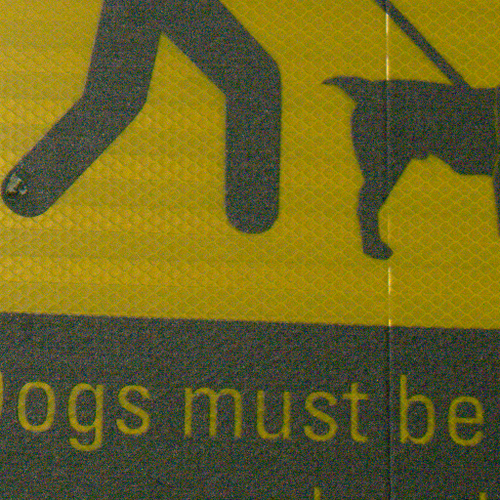}
            \put(78,70){\color{red}\oval(40,50)}
            \end{overpic}  &
            \begin{overpic}[width=0.33\linewidth]{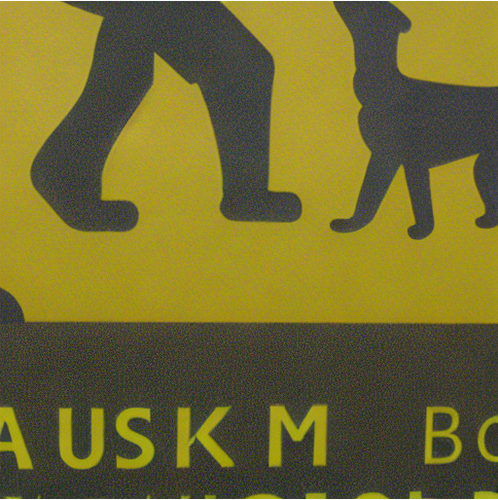}
            \put(78,70){\color{red}\oval(40,50)}
            \end{overpic} 
        \end{tabular}
        \vspace{-0.2cm}
        \caption{Naively using the noisy image as the conditional image in LDM fails to preserve local structures and leads to hallucinations.}
        \label{fig:xattn_flatten}
    \end{minipage}
    \hfill
    \begin{minipage}[t]{0.48\linewidth}
        \centering
        \small
        \begin{tabular}{@{}c@{\hspace{0.5mm}}c@{\hspace{0.5mm}}c}
            Input \& Reference & w/o Residual & w/ Residual (Ours) \\
            \includegraphics[width=0.32\linewidth]{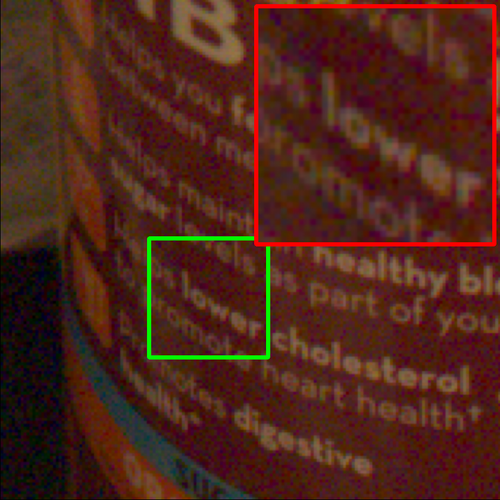} &
            \includegraphics[width=0.32\linewidth]{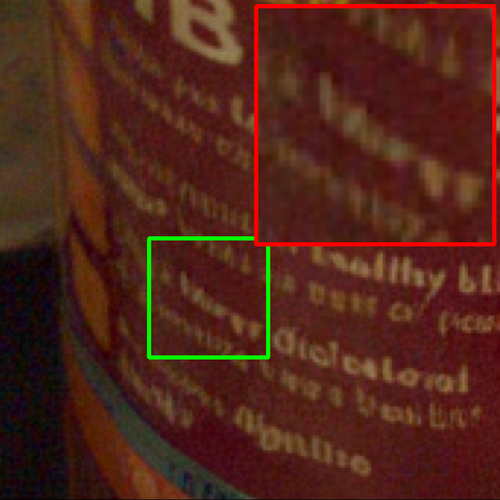} &
            \includegraphics[width=0.32\linewidth]{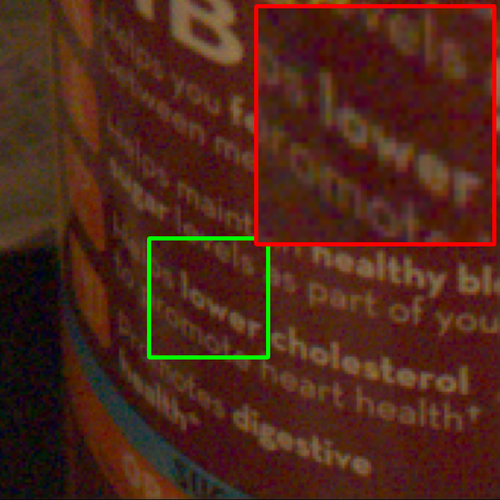}
        \end{tabular}
        \vspace{-0.2cm}
        \caption{VAE reconstruction results with and without our proposed residual architecture. We observe a loss in input details when not using a residual connection.}
        \label{fig:skip_con_gt}
    \end{minipage}
\end{figure*}

\myparagraph{Image restoration with diffusion.}
Diffusion models~\cite{ho2020denoising,sohl2015deep,song2019generative,song2020improved} have been used in image restoration~\cite{wang2022zero,whang2022deblurring,gao2023implicit,lin2023diffbir,ozdenizci2023restoring}. There are also zero-shot methods~\cite{fei2023generative, kawar2022denoising,lugmayr2022repaint,wang2022zero} that utilize the generative power of pre-trained models such as Stable Diffusion~\cite{rombach2022high} without further training. A more comprehensive survey can be found in~\cite{he2024diffusion}. Those methods achieved great success by leveraging diffusion generative models trained from large-scale sRGB data. However, there is no raw image dataset at a similar scale. And the way to bridge the domain gap between raw and sRGB is unknown for those diffusion models.  Naively using the noisy image as the conditional image in a pre-trained Latent Diffusion Model does not preserve the image content, as shown in~\figref{fig:xattn_flatten}.

{\noindent\bf Cross-attention modules.}
Cross-attention module~\cite{chen2021crossvit} was initially designed to compute attention between different token sequences. This idea has been extended to multi-modal tasks, \eg, cross-attention is used to align text and image features to improve text-conditioned image-to-image tasks~\cite{hertz2023prompttoprompt,chen2024training}. 
In image restoration, cross-attention modules were used to provide conditional input, \eg, %
\citet{wang2022restoreformer} uses a cross-attention where the corrupted face image is the query, the high-quality image is the key, and the value is the restored face image. \citet{dang2024ppformer} proposed cross-attention between local \& global features for low-light image enhancement in sRGB space. %
In contrast, our proposed region-based cross-attention method preserves local information by grouping tokens. %

\section{Preliminary}

\myparagraph{Latent diffusion models.}
Diffusion models aim to learn a data distribution that is easy to sample from. To make diffusion models more efficient, Latent Diffusion Models (LDMs)~\cite{rombach2022high} use a Variational Autoencoder (VAE) that transforms the data into a latent space, then models diffusion in this space. 

LDM begins by passing an input $\vx_0$ through the encoder $\gE$, to obtain latent representations $\rvz_0$. The {\it forward diffusion process} gradually corrupts $\rvz_0$ with additive Gaussian noise using the following Markov Chain:
$
    q(\rvz_t | \rvz_{t-1}) = \mathcal{N}(\rvz_t; \sqrt{\alpha_t} \rvz_{t-1}, (1-\alpha_t) \mI),
$
where $\alpha_t$ is the noise schedule and $t$ is the timestep. The {\it reverse diffusion process} learns a denoising network $\epsilon_\theta$ to undo the added noise. LDM minimizes
\bea\label{eq:ldm_loss}
     \gL_{\text{LDM}} = \mathbb{E}_{\gE(\vx), t, \epsilon \sim \mathcal{N}(0, I)} \left[ w_t \| \epsilon_\theta(\rvz_t, t) - \vn \|_2^2 \right],
\eea
where $w_t$ is a weighting factor and $\vn \sim \gN(0,\mI)$ is a noise sampled from standard Gaussian.

With a trained $\epsilon_\theta$, 
LDM performs sampling directly in the latent space. Starting from a pure Gaussian noise sample $\tilde\rvz_T \sim \gN(0,\mI)$, LDM iteratively denoise this sample
\bea
\tilde{\vz}_{t-1} = \frac{1}{\sqrt{\alpha_t}} \left(\tilde\vz_t - \frac{1-\alpha_t}{\sqrt{1-\bar\alpha_t}}\epsilon_\theta(\tilde\vz_t,t)\right) + \sigma_t \vn,
\eea
where $\vn \sim \gN(0,\mI)$. Finally, the decoder $\gD$ maps $\tilde\rvz_0$ back to the input space, producing a generated sample $\tilde{\rvx}_0 $.

\myparagraph{Cross-attention in LDMs.}
The backbone of LDM's $\epsilon_\theta$ follows a standard U-Net~\cite{ronneberger2015u}. To take in the additional input or conditioning $\rvc$, LDM introduces a cross-attention layer to align features from U-Net with the conditioning vector $\vc$.

Given the latent features (tokens) $\rvz_t^l \in \mathbb{R}^{N \times d}$ at the $l$-th layer of U-Net and the condition $\rvc \in \mathbb{R}^{M \times d} $, a cross-attention computes the queries, keys, and values as follows:
\bea
    \rmQ^l = \rvz_t^l (\rmW_{Q}^l)^\top, \quad \rmK^l = \rvc (\rmW_{K}^l)^\top, \quad \rmV^l = \rvc (\rmW_{V}^l)^\top,
\eea
where $\rmW_{Q}^l, \rmW_{K}^l, \rmW_{V}^l \in \mathbb{R}^{d \times d_{k}^l}$ are learned projection matrices for each layer $l$, and $d^l_k$ is the dimensionality of the attention space. Then the latent features $\rvz_t^{l+1}$  are updated by
\bea
    \rvz_t^{l+1} = \mathbf{A}^l \rmV^l, \text{~whereu~}
    \mathbf{A}^l = \text{Softmax} \left( {\rmQ^l (\rmK^l)^{\top}} / {\sqrt{d_k^l}} \right).
\eea  
Standard cross-attention provides a global receptive field of the entire image, which does not preserve the local structures of the conditioning image.

\section{Approach}
The task of low-light raw image enhancement is to reconstruct the clean image $\rvx_0$ given its noisy observation $\rvy $ at \textit{raw} domain. 
This is an ill-posed problem that
can be formulated as modeling a conditional probability distribution $p_\theta(\rvx_0 | \rvy)$ using a diffusion model.
Instead of training from scratch, our method leverages the generative capability of a pre-trained diffusion model and retasks it to the low-light enhancement task. We proposed a new framework with the following components to achieve the goal: %
    
    \noindent\textbullet~{\it Customized diffusion raw image enhancement pipeline} (\secref{sec:pipeline}). We designed a customized pipeline to leverage diffusion models in camera ISP. The raw images are first converted into linear RGB images using standard ISP algorithms to significantly reduce the domain gap with sRGB images.
    
    \textbullet~{\it Region-based cross-attention for conditioning}  (\secref{sec:x-atten}). Naive conditioning in LDM preserves image style but not local structure. As shown in~\figref{fig:xattn_flatten}, standard cross-attention leads to structure hallucinations with image content changed.  To address this, we propose a region-based cross-attention %
    which helps to align the local details from the noisy observation $\vy$. %
    
    \textbullet~{\it Content preservation VAE with residuals} (\secref{sec:skip-vae}). While region-based cross-attention improves the local structure, we need finer control to preserve image content identity, such as text, human faces.   To address this, we propose to improve pixel-level fidelity via a residual VAE. %
    
    \textbullet~{\it Decoder loss for reducing color shift} (\secref{sec:lrgb}). Directly fine-tuning the LDM, via~\equref{eq:ldm_loss}, often leads to color shift, as the supervision is only in the latent space. To address this, we propose a decoder-based reconstruction loss, which provides a supervised loss in the pixel space. %

{\noindent\bf Overview.} 
A visual illustration of our method is provided in~\figref{fig:pipeline}. Given a noisy image $\rvy$, we use the VAE encoder with residuals to get the condition latent $\rvz_\rvy$. This latent is then processed by a context processor, which generates condition features $\rvz_\rvy^l$ for each layer $l$. Simultaneously, we sample a noise $\tilde\rvz_T \sim \gN(0, \mI)$ and pass it through the denoising U-Net encoder to extract latent features $\tilde\rvz_t^l$ for each layer.
To condition the image features on condition features, we employ a region-based cross-attention module. After iterative processing through the network, the final denoised latent $\tilde\rvz_0$ is input into the residual VAE decoder to provide the final denoised SRGB image $\tilde\rvx_0$.

\begin{figure*}[t]
    \centering
    \includegraphics[width=\linewidth]{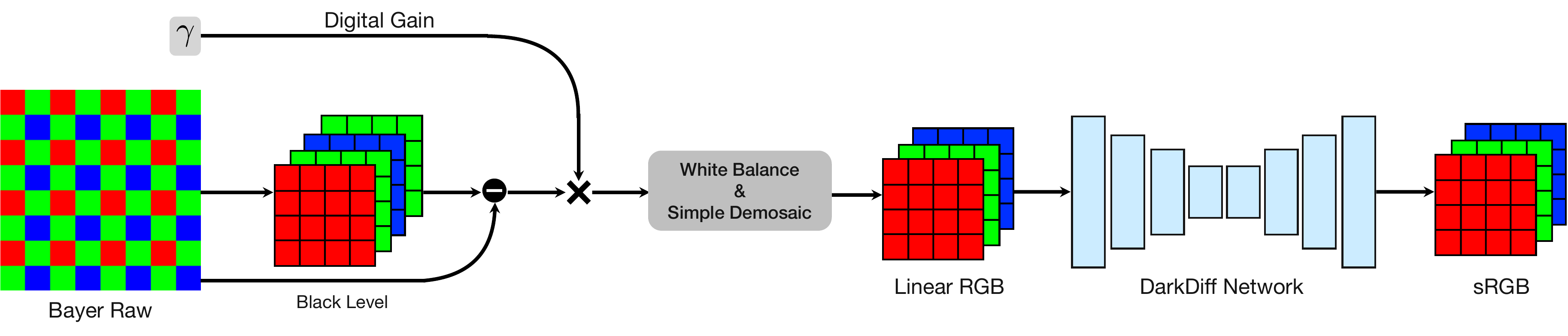}
    \vspace{-0.25cm}
    \caption{Our data processing pipeline converts Bayer raw input into a linear RGB format by applying white balance and demosaicing to the packed and amplified data. This linear RGB image is then passed to the diffusion model to produce the final sRGB image.
    }
    \vspace{-0.25cm}
    \label{fig:data_pipeline}
\end{figure*}

\subsection{Customized diffusion raw image pipeline}
\label{sec:pipeline}
As StableDiffusion was pre-trained on sRGB images and not raw images, we designed this pipeline to reduce this domain gap. Before the enhancement network, we pack Bayer raw data into separate RGBG channels, linearize it by subtracting the black level, and then apply a large digital gain (amplification ratio in SID~\cite{chen2018learning}) to brighten the image. %
We then apply white balance and simple demosaic such that the data is in brightened linear RGB (LRGB) space. The diffusion model only learns partial ISP processing of denoising and maps the noisy linear RGB images to sRGB images; See~\figref{fig:data_pipeline}. With this design, the noise on the original raw images is evenly amplified and still remains zero mean, which are good properties for the enhancement network to deal with.  

\begin{figure*}[t]
    \centering
    \begin{minipage}[t]{0.48\linewidth}
        \centering
        \renewcommand{\arraystretch}{0.2}
        \includegraphics[width=\linewidth]{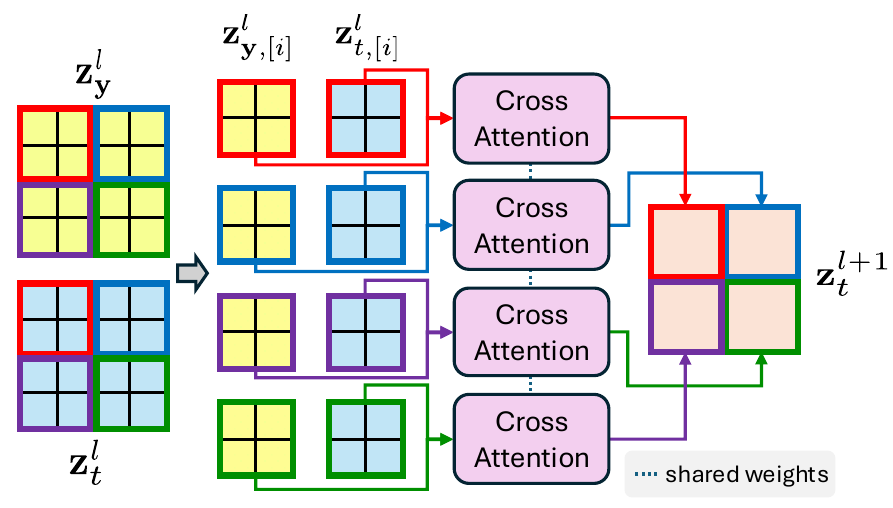}
        \vspace{-0.1cm}
        \caption{Region-based cross-attention. 
        }
        \label{fig:pxattn_pipeline}
    \end{minipage}
    \hfill
    \begin{minipage}[t]{0.48\linewidth}
        \centering
        \includegraphics[width=\linewidth]{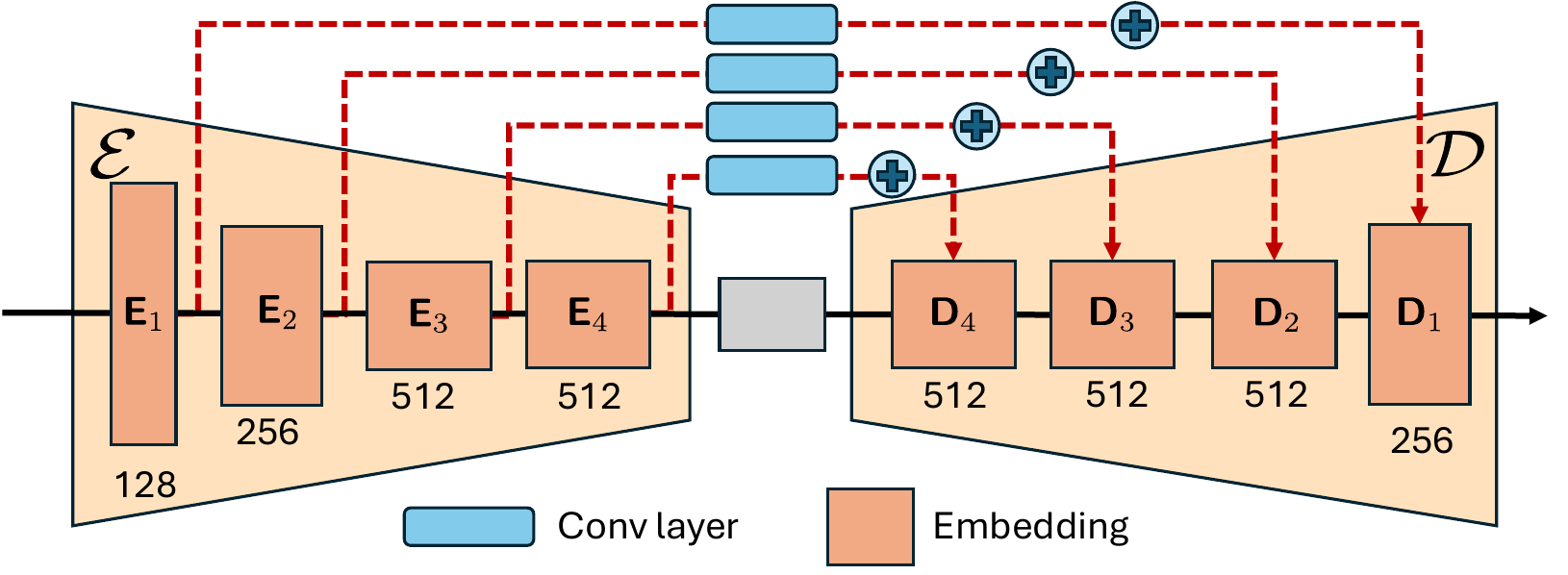}
        \vspace{-0.1cm}
        \caption{Residual VAE architecture.}
        \label{fig:skip_connection_pipeline}
    \end{minipage}
    \vspace{-0.5cm}
\end{figure*}

\subsection{Region-based cross-attention for conditioning}
\label{sec:x-atten}
The main drawback of standard cross-attention is that it lacks locality, \ie, all of the tokens are attentive to \textit{all} the spatial locations in the noisy condition image $\vy$. To introduce a sense of locality to cross-attention, we introduce \textit{region-based cross-attention}, which computes attention over a group of localized image patches. See~\figref{fig:pxattn_pipeline} for illustration.

Formally, the image latent $\rvz_t \in \sR^{H \times W \times d}$ is passed through the encoder of the U-Net, producing the latent feature $\rvz_t^l$ at each cross-attention layer $l$. Similarly, the condition latent $\rvz_\rvy \in \sR^{H \times W \times d}$ is processed by the context processor, which has the same architecture as and weights initialized from the U-Net encoder.
For each cross-attention layer $l$, we group these latents into $K^l$ non-overlapping regions. Each region is then converted into a set of $N'$ tokens. 

Let $\{\rvz_{t, [i]}^{l}\}_{i=1}^{K^l}$ and $\{\rvz_{\rvy, [i]}^{l}\}_{i=1}^{K^l}$ denote the latent and condition region tokens at layer $l$, where $\rvz_{t, [i]}^{l}, \rvz_{\rvy, [i]}^{l} \in \mathbb{R}^{N' \times d}$. The queries, keys, and values for each $i$-th region are:
\bea
    &\rmQ^{l}_{[i]} = \rvz_{t, [i]}^{l} (\rmW_{Q}^l)^\top, \quad \rmK^{l}_{[i]} = \rvz_{\rvy, [i]}^{l} (\rmW_{K}^l)^\top, \quad \\
    &\nonumber \rmV^{l}_{[i]} = \rvz_{\rvy, [i]}^{l} (\rmW_{V}^l)^\top,
\eea
where $\rmW_{Q}^l, \rmW_{K}^l, \rmW_{V}^l$ are \textit{shared} across all the regions. The attention weights for the $i$-th region %
 are computed as:
\begin{equation}
    \rmA_{[i]}^{l} = \text{Softmax} \left( {\rmQ^{l}_{[i]} (\rmK^{l}_{[i]})^\top}/{\sqrt{d_{k}^l}} \right).
\end{equation}
Finally, the updated latent at layer $l$ is computed by attending to the condition latent:
\bea
    \rvz_{t, [i]}^{l+1} = \rmA_{[i]}^{l} \rmV^{l}_{[i]}.
\eea
By operating at the region level and performing cross-attention over multiple layers, the model learns to focus on local structures %
from the condition image. %

\subsection{Content preservation VAE with residuals}
\label{sec:skip-vae}
To extract the latent features, the conditioning image $\vy$ is first processed through an encoder $\gE$ before entering the diffusion process. To preserve the visible content of $\vy$, the encoder must be able to capture the relevant information. As illustrated in~\figref{fig:skip_con_gt}, simply passing an image through the encoder and then the decoder of a standard VAE architecture leads to noticeable pixel differences in the reconstruction.

To address this issue, we introduce residual connections between the corresponding encoder and decoder blocks within the VAE. Formally, denote the image embedding at each block $b$ of the encoder as $\tE_b$. The embedding is passed through a convolution layer $\text{Conv}_b$ such that the channel size of $\text{Conv}_b(\mE_b)$ matches the corresponding decoder embedding $\tD_b$. The residual connection for each block $b$ is expressed as:
$\tD_{b+1} = \tD_b + \text{Conv}_b(\tE_b).$
This architecture encourages the details from the conditioning image $\vy$ to be better preserved across different resolutions, only compressing the necessary information at the VAE's bottleneck. The VAE architecture is shown in~\figref{fig:skip_connection_pipeline}. Empirically, we observe that the proposed architecture significantly improves reconstruction quality.

\subsection{Decoder loss for reducing color shift}
\label{sec:lrgb}
A standard LDM training with~\equref{eq:ldm_loss} only provides supervision in the latent space and suffers from the issue of color shift when the input linear RGB image is too noisy. To ensure that the reconstructed image is visually consistent with the clean ground-truth sRGB image $\rvx_0$, we introduce a loss function on the image space by using the decoder. A multi-step approximation requires backpropagation through all unrolled steps, which is computationally expensive. Instead, we construct a ``one-step'' approximation of the denoised latent following the derivation by~\citet{ho2020denoising}:
\bea \label{eq:latent_approx}
    \hat{\rvz}_0 = \frac{1}{\sqrt{\bar{\alpha}_t}} \left( \rvz_t - \sqrt{1 - \bar{\alpha}_t} \hat{\epsilon}_t \right),
\eea
where $\bar{\alpha}_t = \prod_{s=1}^t \alpha_s$ and $\alpha_t$ is the noise scheduling factor. 

The VAE decoder $\gD$ then maps this approximated latent $\hat{\rvz}_0$ back to the sRGB image space. The image-space reconstruction loss is given by:
\bea \label{eq:image_loss}
    \mathcal{L}_{\text{image}} = \mathbb{E}_{\rvx} \left[ \|\rvx - \gD(\hat{\rvz}_0)\|_2^2 \right].
\eea
As the loss in the sRGB space, it will directly penalize the model if there is a color shift.

\begin{table*}[t]
\begin{minipage}{.31\linewidth}
        \centering
        \caption{Quantitative results on Sony set of the SID dataset~\cite{chen2018learning}, ratio $300$ subset. }
    \resizebox{\linewidth}{!}{%
    \begin{tabular}{lccc}
        \specialrule{.15em}{.05em}{.05em}
        Methods & PSNR$\uparrow$ & SSIM$\uparrow$ & LPIPS$\downarrow$ \\
        \hline
        SID & 27.22 & 0.664 & 0.235 \\
        ELD & 26.69 & 0.659 & 0.225 \\
        LED & 22.14 & 0.5200 & 0.322 \\
        LRD & 27.13 & 0.656 & 0.232 \\
        ED & \bf 27.74 & \bf 0.679 & 0.240 \\
        SD Concat & 17.33 & 0.334 & 0.240 \\
        SD ControlNet & 16.70 & 0.304 & 0.259 \\
        \ours & 26.78 & 0.644 & \bf 0.186 \\
        \specialrule{.15em}{.05em}{.05em}
    \end{tabular}
    }
        \vspace{-0.2cm}
    \label{tab:sid}

\end{minipage}
\hspace{6pt}
\begin{minipage}{.31\linewidth}
        \centering
        \caption{Quantitative results on ELD dataset~\cite{wei2020physics}, ratio $200$ subset.}

    \resizebox{\linewidth}{!}{%
    \begin{tabular}{lccc}
        \specialrule{.15em}{.05em}{.05em}
        Methods & PSNR$\uparrow$ & SSIM$\uparrow$ & LPIPS$\downarrow$ \\
        \hline
        SID & 27.14 & 0.833 & 0.175 \\
        ELD & 27.14 & 0.832 & 0.159 \\
        LED & 24.50 & 0.674 & 0.210 \\
        LRD & \bf 29.22 & 0.832  & 0.197 \\
        ED & 28.17 & \bf 0.863 & 0.159 \\
        SD Concat & 21.17 & 0.334 & 0.279 \\
        SD ControlNet & 21.89 & 0.422 & 0.234 \\
        \ours & 27.06 & 0.837 & \bf 0.150  \\
        \specialrule{.15em}{.05em}{.05em}
    \end{tabular}
    }
    \vspace{-0.18cm}
    \label{tab:eld}

\end{minipage}
\hspace{6pt}
\begin{minipage}{.31\linewidth}
    \setlength{\tabcolsep}{4pt}
    \centering
     \caption{Quantitative results on LRD dataset~\cite{zhang2023towards}, -3EV subset. The existing methods with high PSNR/SSIM tend to oversmooth the images and perform worse on the perceptual metric LPIPS. }
    \resizebox{\linewidth}{!}{%
    \begin{tabular}{lccc}
        \specialrule{.15em}{.05em}{.05em}
        Methods & PSNR$\uparrow$ & SSIM$\uparrow$ & LPIPS$\downarrow$ \\
        \hline
        LRD & \bf 30.69 &\bf 0.802 & 0.155 \\
        SD Concat & 18.99 & 0.475 & 0.179 \\
        SD ControlNet & 17.83 & 0.444 & 0.190 \\
        \ours & 29.45 & 0.776 & \bf 0.103 \\
        \specialrule{.15em}{.05em}{.05em}
    \end{tabular}
    }
        \vspace{-0.2cm}
    \label{tab:lrd}

\end{minipage}
\vspace{-0.3cm}
\end{table*}

\subsection{Training and implementation details}

{\bf Training details.} We train the proposed model in two stages. We first fine-tune the pre-trained VAE on pairs of noisy linear RGB and clean sRGB images following the loss function proposed by~\citet{rombach2022high}, which is a combination of the perceptual loss~\cite{zhang2018unreasonable}, a patch-based adversarial loss~\cite{esser2021taming}, and a KL divergence regularizing the latent space. %

In the second stage, we train the latent diffusion model using the latent representations produced by the fine-tuned VAE. The U-Net's weights are initialized using the weights from a pre-trained LDM. The training objective is given by
$%
\gL = \gL_{\text{LDM}} + \lambda \cdot \gL_{\text{image}},
$ %
where $\lambda$ balances between the two loss functions. More implementation details are in the Appendix.

\myparagraph{Controlling generation with classifier-free guidance.} 
During inference time, we control the level of generation in DarkDiff by using classifier-free guidance~\cite{ho2022classifier} and extending the denoiser
\begin{wrapfigure}{r}{0.5\textwidth}
    \centering
       \vspace{-0.2cm}
        \footnotesize
        \begin{tabular}{c@{\hspace{1mm}}c@{\hspace{1mm}}c}
            Noisy Input & guidance=1 & guidance=5 \\
            \includegraphics[width=0.3\linewidth]{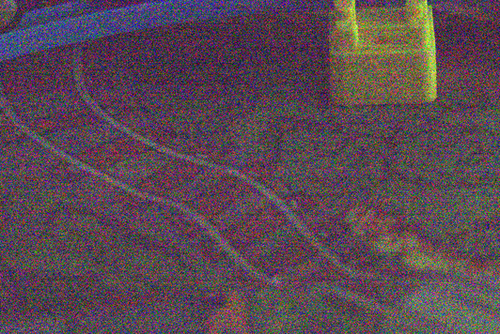} &
            \includegraphics[width=0.3\linewidth]{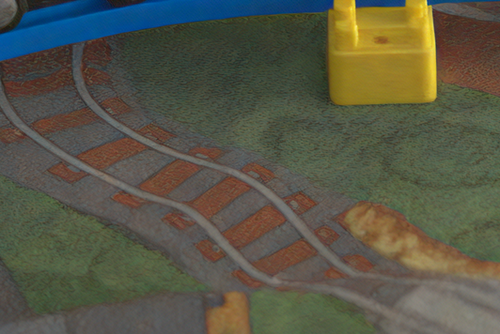} &
            \includegraphics[width=0.3\linewidth]{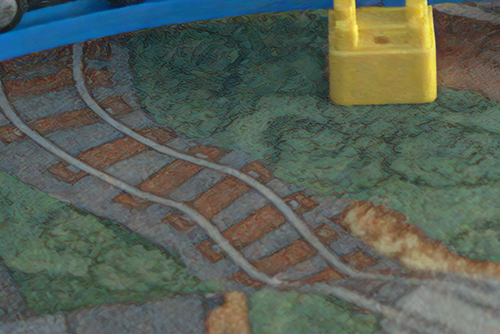} \\
            \includegraphics[width=0.3\linewidth]{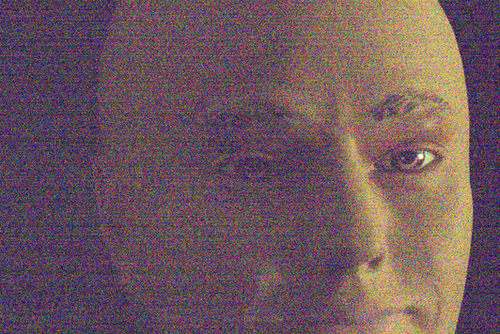} &
            \includegraphics[width=0.3\linewidth]{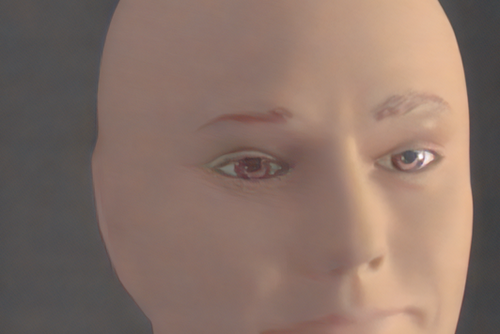} &
            \includegraphics[width=0.3\linewidth]{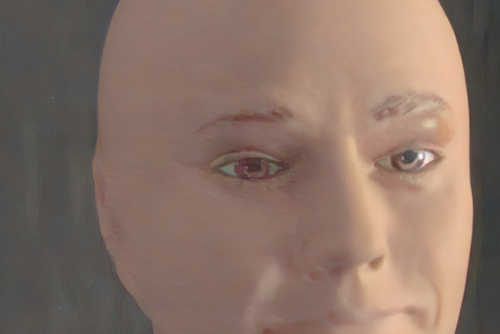} \\
        \end{tabular}
        \caption{Results of classifier-free guidance at different strengths. A small amount of guidance smooths patterns; strong guidance adds details.}
        \label{fig:classfier_free_guidance}
\end{wrapfigure}
to take in an additional input, \ie, $\epsilon_\theta(\vz_t, \vc, t)$. The adjusted noise prediction $\hat{\epsilon}$ is controlled by linearly combining the conditional and unconditional predictions, \ie,
\bea
\label{eq:cfg}
\hat{\epsilon} = \epsilon_\theta(\tilde\vz_t, t) + \omega \cdot \left( \epsilon_\theta(\tilde\vz_t, \vc, t) - \epsilon_\theta(\tilde\vz_t, t) \right),
\eea
where $\epsilon$ is the guidance weight.
As shown in~\figref{fig:classfier_free_guidance}, when guidance is one, the model tends to smooth the details of the ground and background with less generation, \eg, on the texture of the cloth. Larger guidance helps the model generate details while also having the potential to introduce artifacts and more hallucinations.

\section{Experiments}
\myparagraph{Datasets.}
We evaluate our proposed framework on three widely used low-light raw image datasets:\hspace{-0.4cm}
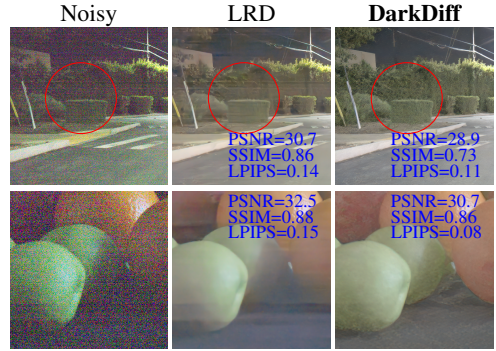
\begin{wrapfigure}[17]{r}{0.5\textwidth}
    \centering
    \vspace{-9pt}
    \small
    \setlength{\tabcolsep}{1pt}
    \begin{tabular}{ccc}
    Noisy & LRD & \ours \\
    \begin{overpic}[width=0.15\textwidth]{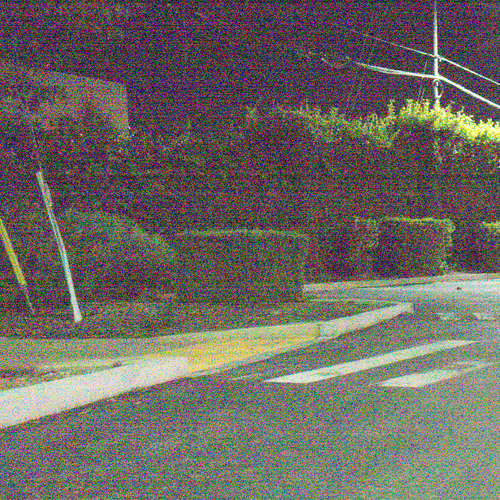}
        \put(45,55){\color{red}\circle{45}} 
    \end{overpic} &
    \begin{overpic}[width=0.15\textwidth]{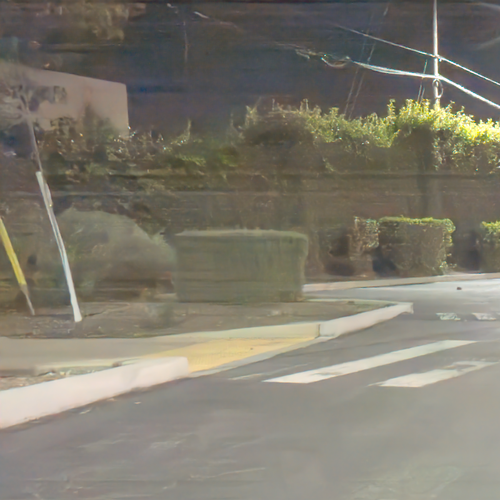}
        \put(35,25){\color{blue}\scriptsize PSNR=30.7}
        \put(35,15){\color{blue}\scriptsize SSIM=0.86}
        \put(35,5){\color{blue}\scriptsize LPIPS=0.14}
        \put(45,55){\color{red}\circle{45}} 
    \end{overpic} &
    \begin{overpic}[width=0.15\textwidth]{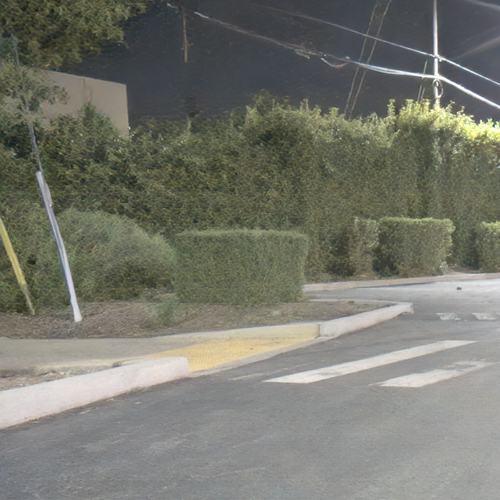}
        \put(35,25){\color{blue}\scriptsize PSNR=28.9}
        \put(35,15){\color{blue}\scriptsize SSIM=0.73}
        \put(35,5){\color{blue}\scriptsize LPIPS=0.11}
        \put(45,55){\color{red}\circle{45}} 
    \end{overpic} \\
    \begin{overpic}[width=0.15\textwidth]{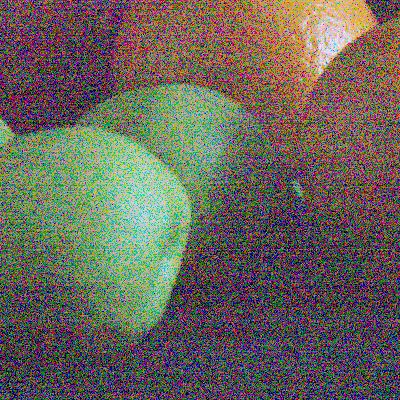}
    \end{overpic} &
    \begin{overpic}[width=0.15\textwidth]{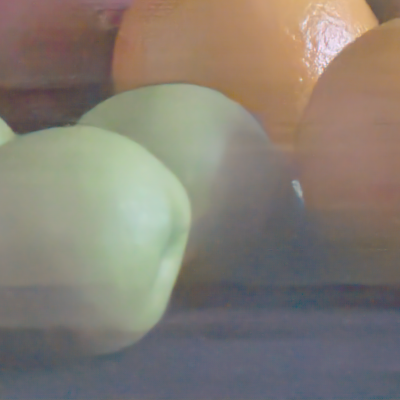}
        \put(35,90){\color{blue}\scriptsize PSNR=32.5}
        \put(35,80){\color{blue}\scriptsize SSIM=0.88}
        \put(35,70){\color{blue}\scriptsize LPIPS=0.15}
    \end{overpic} &
    \begin{overpic}[width=0.15\textwidth]{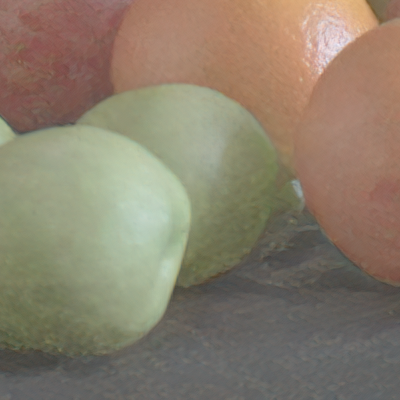}
        \put(35,90){\color{blue}\scriptsize PSNR=30.7}
        \put(35,80){\color{blue}\scriptsize SSIM=0.86}
        \put(35,70){\color{blue}\scriptsize LPIPS=0.08}
    \end{overpic}
    \end{tabular}
    \vspace{-6pt}
    \caption{LPIPS better reflects perceptual quality than PSNR or SSIM~\cite{zhang2018unreasonable}.}
    \label{fig:metric_human}
\end{wrapfigure}

SID~\cite{chen2018learning}, ELD~\cite{wei2020physics}, and the recently introduced LRD~\cite{zhang2023towards}. 
For the SID dataset, we follow the split used by \citet{wei2020physics}, utilizing the Sony subset and training our models for both SID and ELD evaluations on the 280 image pairs in the SID training set. For the LRD test set, we train the model using the corresponding LRD training set, which contains 1,440 training pairs. 
Images captured in bright light normally have high SNR values and do not necessarily require content generation. We focus on the most challenging cases: extreme low-light images with high digital gain: $\times$300 in SID, $\times$200 in ELD, and -3EV in LRD. For all experiments, we use the real captured paired data for evaluation. 

\myparagraph{Evaluation metrics.} 
We evaluate the performance of our framework using the Peak Signal-to-Noise Ratio (PSNR), Structural Similarity Index (SSIM), and Learned Perceptual Image Patch Similarity (LPIPS). While PSNR and SSIM are commonly reported metrics, they {\bf do not always correlate well with human perception for generative tasks}~\cite{gu2020pipal,Gu_2022_CVPR,blau2018perception}; nonetheless, we report them for completeness. 
For image-to-image tasks, it is more common to use LPIPS to evaluate the generative capability~\cite{karras2020analyzing,kawar2023imagic,couairon2023diffedit}. 
We illustrate the behavior of these metrics in~\figref{fig:metric_human}.

\begin{table*}[t]
\centering
\begin{minipage}{.32\linewidth}
\setlength{\tabcolsep}{5pt}
    \centering
    \caption{Ablation results of the contribution of each component in the model framework.} %
    \resizebox{\columnwidth}{!}{%
    \begin{tabular}{ccc|ccc}
        \specialrule{.15em}{.05em}{.05em}
        \multicolumn{3}{c|}{\textbf{Components}} & \multicolumn{3}{c}{\textbf{Metrics}} \\
        \cmidrule(lr){1-3} \cmidrule(lr){4-6}
        \S\ref{sec:x-atten} & \S\ref{sec:skip-vae} & \S\ref{sec:lrgb} & {PSNR} & {SSIM} & {LPIPS} \\
        \hline
         & \checkmark & \checkmark &  26.52 & 0.642 & 0.198\\
         \checkmark & & \checkmark & 22.44 & 0.566 &0.224 \\
         \checkmark & \checkmark &  & 26.40 & 0.642 & 0.198 \\
        \hline
        \checkmark & \checkmark & \checkmark & \bf 26.78 & \bf 0.644 & \bf 0.186 \\
        \specialrule{.15em}{.05em}{.05em}
    \end{tabular}
    }
    \vspace{-0.15cm}
    \label{tab:ablation_components}

\end{minipage}
\hspace{3pt}
\begin{minipage}{.31\linewidth}
\setlength{\tabcolsep}{4pt}
    \centering
    \small
     \caption{Ablation on the VAE's input and output format. The proposed LRGB-sRGB performs the best.}
    \resizebox{\linewidth}{!}{%
    \begin{tabular}{lccc}
        \specialrule{.15em}{.05em}{.05em}
        Data & PSNR & SSIM & LPIPS \\
        \hline
        Raw-Raw & 22.17 & 0.451 & 0.265 \\
        Raw-sRGB & 25.34 & 0.614 & 0.204 \\
        LRGB-sRGB & \bf 26.78 & \bf 0.644 & \bf 0.186 \\
        \specialrule{.15em}{.05em}{.05em}
    \end{tabular}
    }
        \vspace{-0.15cm}
    \label{tab:ablation_input_output}

\end{minipage}
\hspace{3pt}
\begin{minipage}{.31\linewidth}
\setlength{\tabcolsep}{5pt}
    \centering
    \small
    \caption{Ablation results of image conditioning modules. All the results here are {\it using our proposed residual VAE}.}
    \resizebox{\linewidth}{!}{%
    \begin{tabular}{lccc}
        \specialrule{.15em}{.05em}{.05em}
        Module & PSNR & SSIM & LPIPS \\
        \hline
        Concat &  26.52 & 0.644 & 0.192 \\
        ControlNet & 26.54 & 0.642 & 0.211 \\
        Global XAttn & 21.00 & 0.541 & 0.381 \\
        Rb-XAttn & \bf 26.78 & \bf 0.644 & \bf 0.186  \\
        \specialrule{.15em}{.05em}{.05em}
    \end{tabular}
    }
        \vspace{-0.2cm}
    \label{tab:ablation_condition_module}

\end{minipage}
\vspace{-0.1cm}
\end{table*}

{\noindent\bf Baselines.} We consider the following groups of baselines.

\begin{itemize}[topsep=0pt, leftmargin=16pt,nosep]
    \setlength\itemsep{0em}
    \setlength{\parskip}{2.5pt}
    \item \textit{Regression-based}: SID~\cite{chen2018learning}, ELD~\cite{wei2020physics}, LED~\cite{jiniccv23led}, and LRD~\cite{zhang2023towards}.
    \item \textit{Diffusion-based}: ExposureDiffusion (ED)~\cite{wang2023exposurediffusion}, SD Concat, and SD ControlNet.
{\it To the best of our knowledge, ED is the current SOTA on public benchmarks.}
\end{itemize}

For {\it SD Concat}, we modified the model by concatenating the noisy latent $\rvz_\rvy$ as input to the diffusion process. This is inspired by~\citet{brooks2023instructpix2pix}, where they use concat to condition a diffusion model for the task of image editing. 
{\it SD ControlNet} is a baseline using ControlNet~\cite{zhang2023adding} as the conditioning module. For all methods as well as ground-truth, we use the same ISP process,~\ie, rawpy library, for final evaluation in sRGB. See the Appendix for more details.

\begin{figure*}[t]
    \centering
    \small
    \resizebox{1\textwidth}{!}{%
    \begin{tabular}{c@{\hspace{1mm}}c@{\hspace{1mm}}c@{\hspace{1mm}}c@{\hspace{1mm}}c}
        Noisy Input & Reference & LRD & Exposure Diffusion & \ours  \\

        \includegraphics[width=0.19\linewidth]{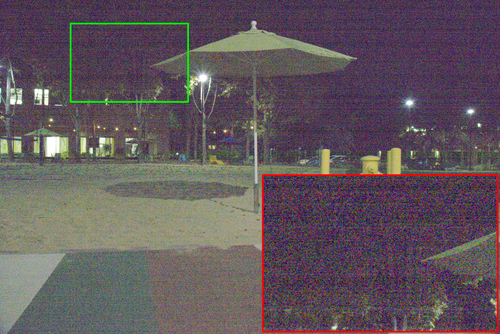} & 
        \includegraphics[width=0.19\linewidth]{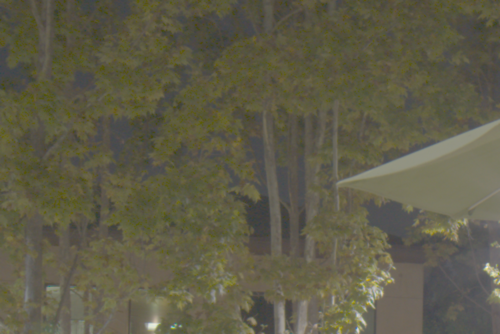} &
        \includegraphics[width=0.19\linewidth]{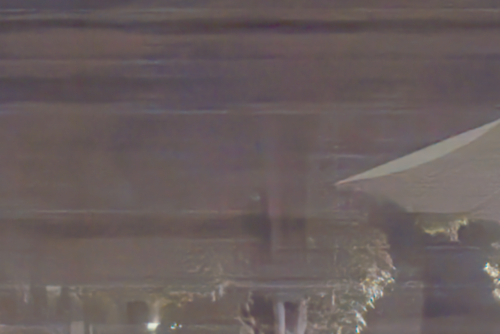} &
        \includegraphics[width=0.19\linewidth]{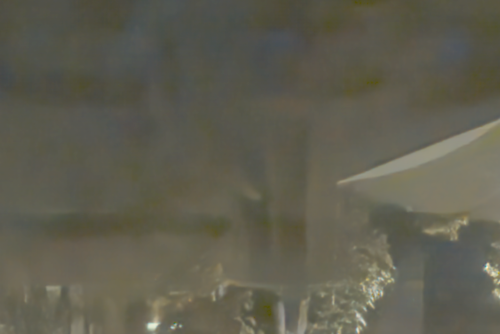} &
        \includegraphics[width=0.19\linewidth]{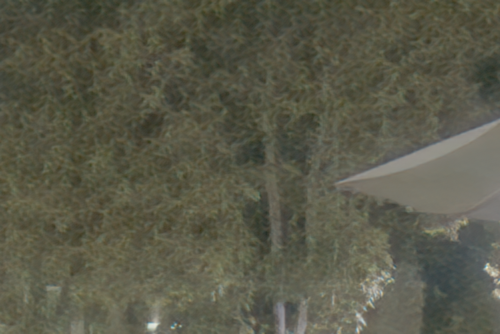} \\
        
        \includegraphics[width=0.19\linewidth]{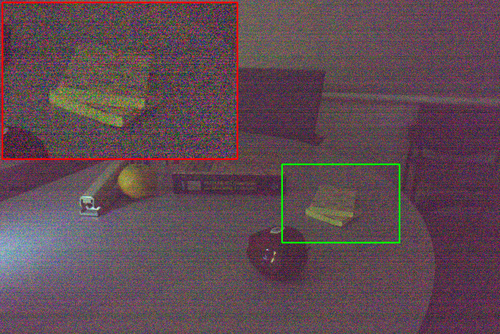} & 
        \includegraphics[width=0.19\linewidth]{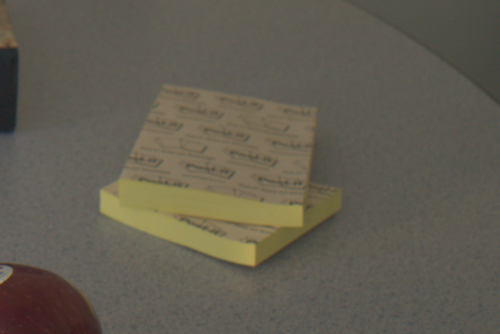} &
        \includegraphics[width=0.19\linewidth]{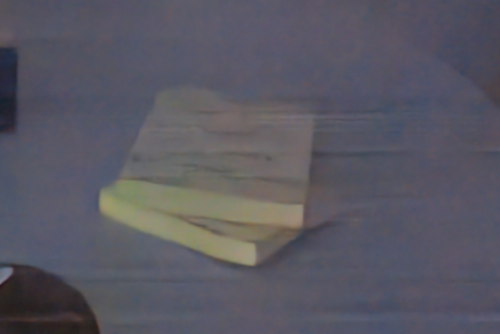} &
        \includegraphics[width=0.19\linewidth]{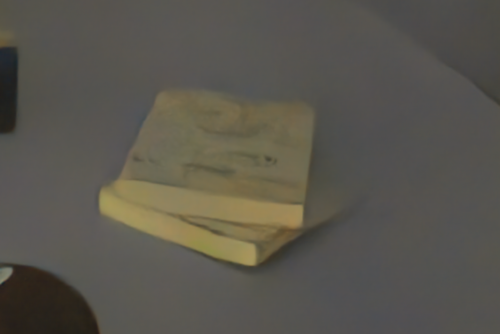} &
        \includegraphics[width=0.19\linewidth]{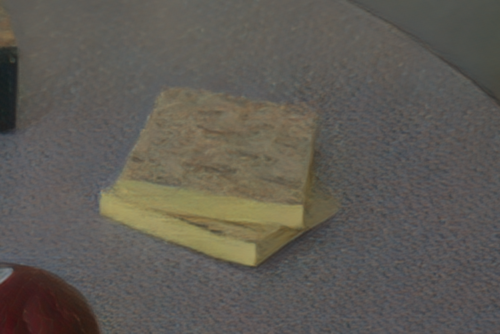} \\

        \includegraphics[width=0.19\linewidth]{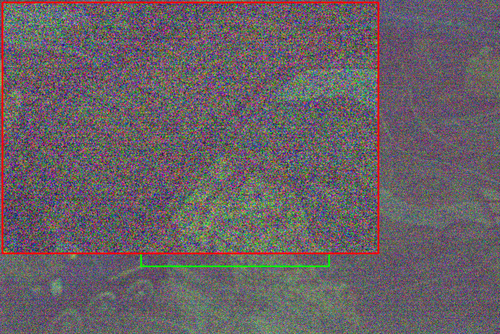} & 
        \includegraphics[width=0.19\linewidth]{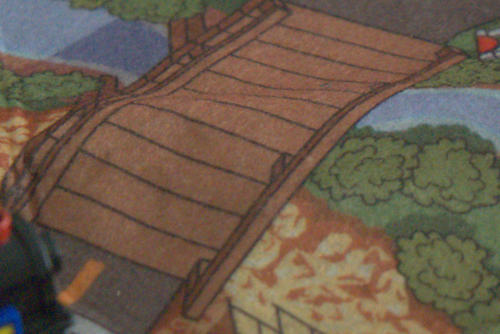} &
        \includegraphics[width=0.19\linewidth]{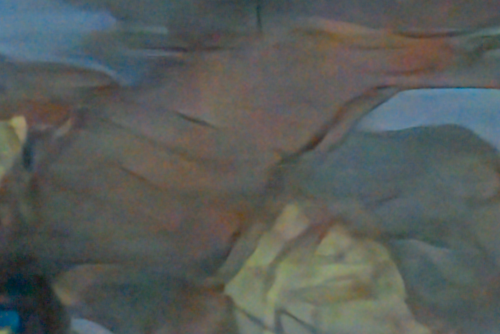} &
        \includegraphics[width=0.19\linewidth]{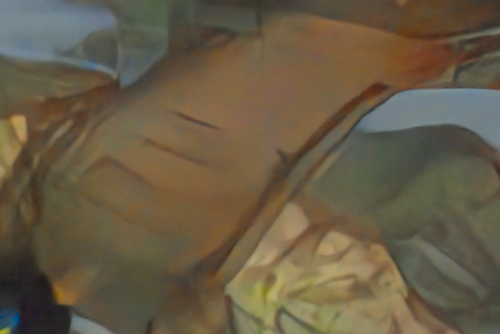} &
        \includegraphics[width=0.19\linewidth]{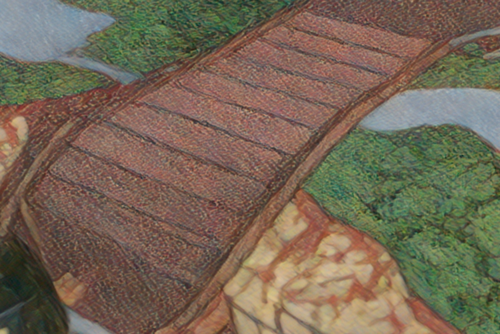} \\

    \end{tabular}
    }
    \vspace{-0.25cm}
    \caption{Low-light image denoising results on SID~\cite{chen2018learning}. Our method generates patterns on the trees, the sticky notes, and the bridge. In contrast, the baselines' results are smoothed out, \eg, the details of the trees are barely visible. 
    }
    \vspace{-.4cm}
    \label{fig:compare_w_baseline}
\end{figure*}

\begin{figure*}[t]
    \centering
    \small
    \resizebox{\linewidth}{!}{%
    \begin{tabular}{c@{\hspace{1mm}}c@{\hspace{1mm}}c@{\hspace{1mm}}c@{\hspace{1mm}}c}
        Noisy Input & Reference & LRD & Exposure Diffusion & \ours  \\

        \includegraphics[width=0.244\linewidth]{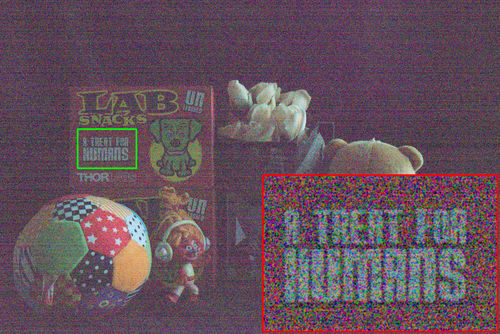} &
        \includegraphics[width=0.244\linewidth]{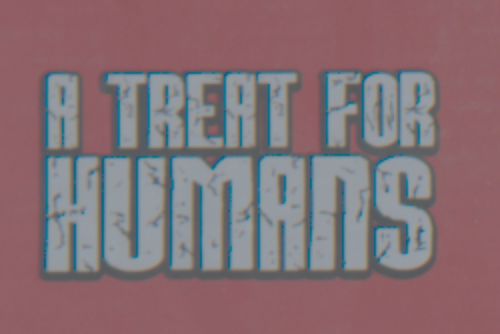} &
        \includegraphics[width=0.244\linewidth]{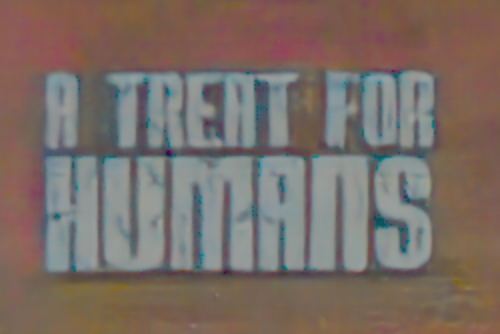} &
        \includegraphics[width=0.244\linewidth]{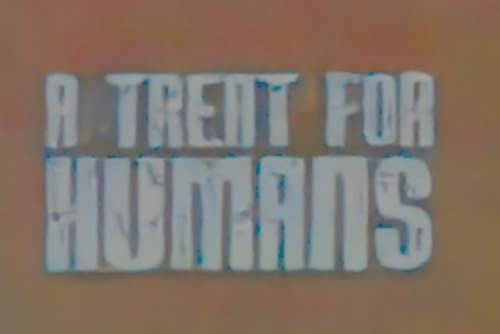} &
        \includegraphics[width=0.244\linewidth]{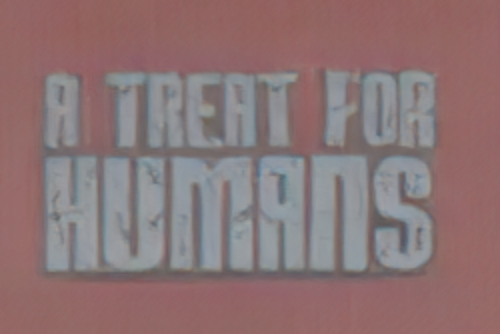} 
        \\

         \includegraphics[width=0.244\linewidth]{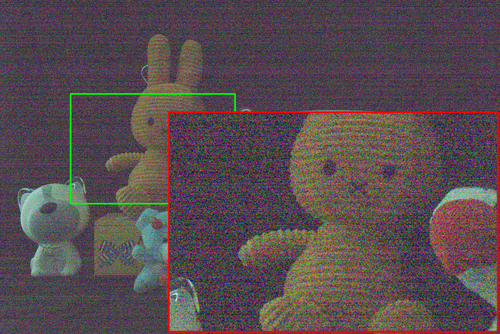} &
        \includegraphics[width=0.244\linewidth]{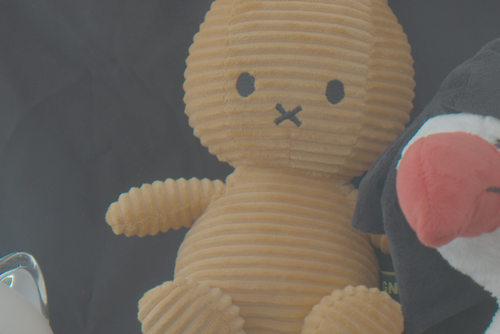} &
        \includegraphics[width=0.244\linewidth]{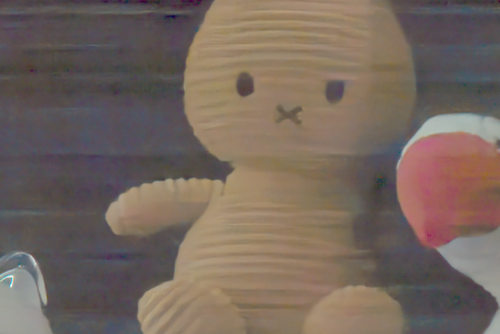} &
        \includegraphics[width=0.244\linewidth]{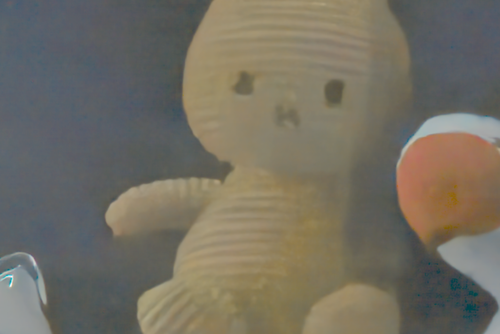} &
        \includegraphics[width=0.244\linewidth]{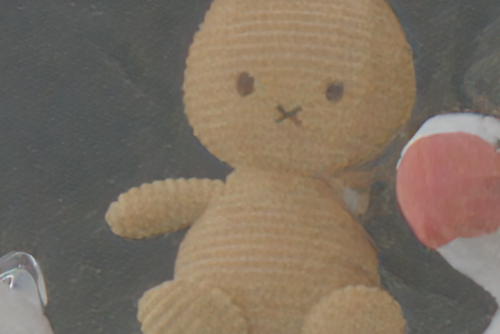} 
        \\
  
    \end{tabular}
    }
    \vspace{-0.25cm}
    \caption{Low-light image denoising results on ELD~\cite{wei2020physics}. Our results maintain the color of the images and have fewer artifacts than the baseline results. Notably, the furry texture of the brown plush is more vivid in our result.
    }
    \vspace{-.4cm}
    \label{fig:main_qualitative_eld}
\end{figure*}

\begin{figure*}[t]
\small
    \centering
    \begin{tabular}{c@{\hspace{1mm}}c@{\hspace{1mm}}c@{\hspace{1mm}}c@{\hspace{1mm}}c}
        Noisy Input & Reference  & LRD & \ours  \\

        \includegraphics[width=0.24\linewidth]{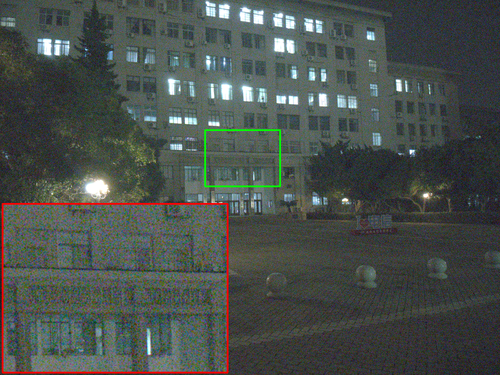}& 
        \includegraphics[width=0.24\linewidth]{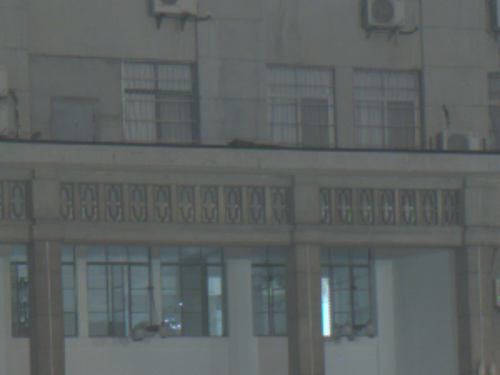} &
        \includegraphics[width=0.24\linewidth]{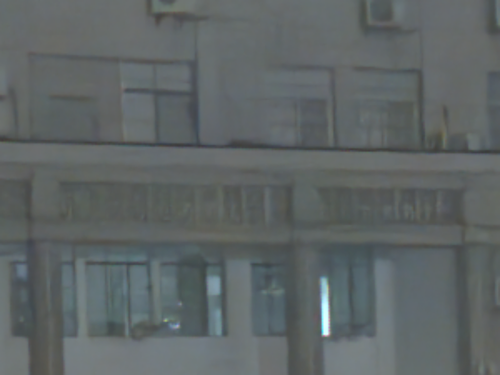} &
        \includegraphics[width=0.24\linewidth]{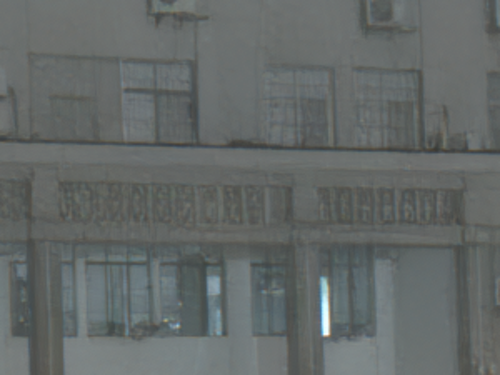} \\

        \includegraphics[width=0.24\linewidth]{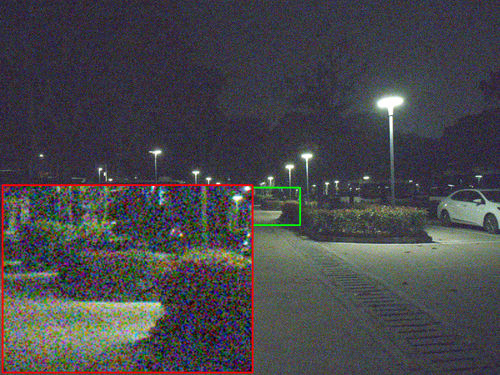} & 
        \includegraphics[width=0.24\linewidth]{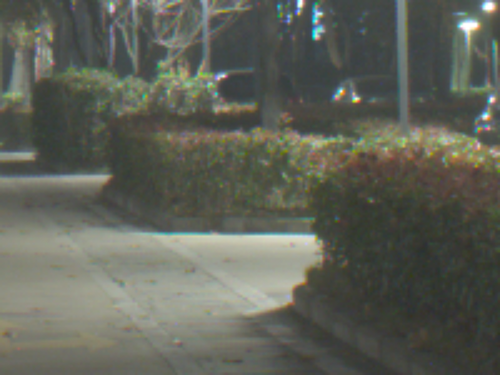} &
        \includegraphics[width=0.24\linewidth]{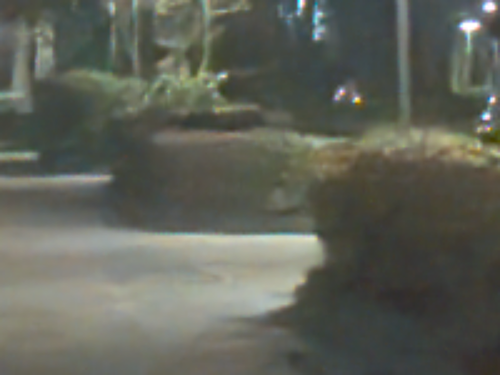} &
        \includegraphics[width=0.24\linewidth]{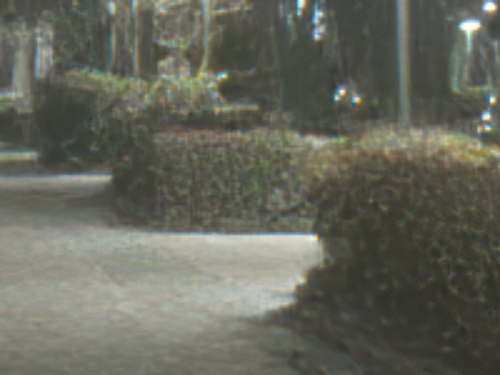} \\        
    \end{tabular}
    \vspace{-0.2cm}
    \caption{Low-light image denoising results on LRD~\cite{zhang2023towards}. Our method generates the details of the building pattern and the textures of the bushes. While the SOTA method, LRD, generates overly smoothed images and lacks details. Best viewed on screens instead of printed. 
    }
    \vspace{-0.3cm}
    \label{fig:main_qualitative_lrd}
\end{figure*}

\subsection{Results}

\myparagraph{Quantitative results.}
We report the results on SID~\cite{chen2018learning} dataset in~\tabref{tab:sid}. As we can observe, our method achieves the best LPIPS, outperforming the second-best baseline of ELD by $20.9\%$. Our method remains competitive on PSNR and SSIM.
To understand why PSNR and SSIM are not the best metrics for generative tasks, please see comparisons in~\figref{fig:metric_human}. Our results have more vivid details like tree leaves and sharper textures of the fruits and cloth when compared to LRD. This higher quality is accurately captured by LPIPS but not reflected in PSNR and SSIM. Notice the line artifacts in LRD, as highlighted in the red circle. These limitations of PSNR and SSIM have also been discussed in prior works on ill-posed image restoration~\cite{gu2020pipal,Gu_2022_CVPR,blau2018perception}.

Next, the results on the ELD~\cite{wei2020physics} and LRD~\cite{zhang2023towards} datasets are provided in~\tabref{tab:eld} and~\tabref{tab:lrd}. Similar to the results for SID, we achieve the highest LPIPS while maintaining competitive PSNR and SSIM.

\myparagraph{Qualitative results} on SID~\cite{chen2018learning} is shown in~\figref{fig:compare_w_baseline}. We observe that in regions with large noise (extremely dark areas), our method reconstructs realistic details, \eg, trees in the background and texture on sticky notes. %
In comparison, the state-of-the-art methods failed to recover enough image details in those most challenging cases.

For the ELD dataset~\cite{ronneberger2015u}, qualitative results are shown in~\figref{fig:main_qualitative_eld}. We observe that the color of the red box is maintained by our method. The SOTA methods contain visible artifacts in those extremely low SNR regions, including color shift, line artifacts, and over-smoothing.

For the LRD dataset~\cite{zhang2023towards}, despite being captured by a different camera sensor, our method still generalizes well. As shown in~\figref{fig:main_qualitative_lrd}, our method consistently generates more detailed images than the state-of-the-art method, avoiding an overly smoothed image.
More qualitative results are provided in the appendix.

\subsection{Ablation Studies}
We conduct ablation studies using the SID dataset~\cite{chen2018learning}.

\myparagraph{Ablating model components.} 
We investigate the importance of various components introduced by our model, including region-based cross-attention (\secref{sec:x-atten}), residual VAE (\secref{sec:skip-vae}), and decoder-based reconstruction loss. (\secref{sec:lrgb}). Here, we replace region-based cross-attention with concatenation in the diffusion input in the first row. As shown in~\tabref{tab:ablation_components}, the performance drops if any components are removed. The result also shows that residual VAE plays a crucial role in the performance gain.

\myparagraph{Impact of data format.}
We evaluate the effect of different input/output data formats on low-light imaging. Since the baseline methods utilize either raw or sRGB formats, we compare our approach with these in~\tabref{tab:ablation_input_output} and demonstrate that using linear RGB input aligns more effectively with the pre-trained model's input spaces while also preserving the information of the raw data.

\myparagraph{Ablating conditioning modules.}
To study the effectiveness of region-based cross attention, %
we compare it with other alternative conditioning methods, including direct concatenation (Concat), ControlNet, and global cross attention. This ablation is conducted using the proposed residual VAE trained in stage 1, \ie, we are only ablation the conditioning module in stage 2 training. As shown in~\tabref{tab:ablation_condition_module}, the proposed region-based attention mechanism achieves the highest quality, while Concat and ControlNet perform similarly.

\section{Conclusion}
We proposed a novel approach to enhancing raw images taken in extreme low-light conditions. We retasked the generative capability of a pre-trained diffusion model into Camera ISP by carefully considering which part of the ISP should be replaced with a diffusion model. %
Our method improves common issues in extreme low-light image enhancement of over-smoothing, local structural fidelity, and color shifts. We propose a region-based cross-attention, a content-preserving VAE, and a decoder-based reconstruction loss to better leverage the diffusion model over the generated content. Experiments on %
three datasets demonstrate that our approach outperforms SOTA methods in perceptual quality.

\clearpage
{
    \small
    \bibliographystyle{ieeenat_fullname}
    \bibliography{main}
}

\clearpage

\newcommand{\beginsupplementary}{%
    \setcounter{section}{0}
	\renewcommand{\thesection}{A\arabic{section}}
	\renewcommand{\thesubsection}{\thesection.\arabic{subsection}}

	\renewcommand{\thetable}{A\arabic{table}}%
	\setcounter{table}{0}

	\renewcommand{\thefigure}{A\arabic{figure}}%
	\setcounter{figure}{0}
	
	\setcounter{algorithm}{0}
}
\beginsupplementary

\onecolumn

\vspace{0.1cm}

{\noindent\Large \bf Appendix\\}

\noindent The appendix is organized as follows:
\begin{itemize}[topsep=0pt, leftmargin=16pt]
    \setlength{\itemsep}{0.0pt}
    \setlength{\parskip}{2.5pt}
    \item In~\secref{sec:qual_results}, we provide additional qualitative results. Please also see the attached supplementary material for more comparisons.
    \item In~\secref{sec:implemtation_details}, we provide additional experimental details in training, inference, and baselines.
    \item In~\secref{sec:limitations}, we illustrate the limitations of our method.
\end{itemize}

\section{Additional qualitative results.}
\label{sec:qual_results}
\begin{figure*}[t]
    \centering
    \begin{tabular}{c@{\hspace{1mm}}c@{\hspace{1mm}}c}
        Noisy Input & \ours & SD Concat \\

        \includegraphics[width=0.323\linewidth]{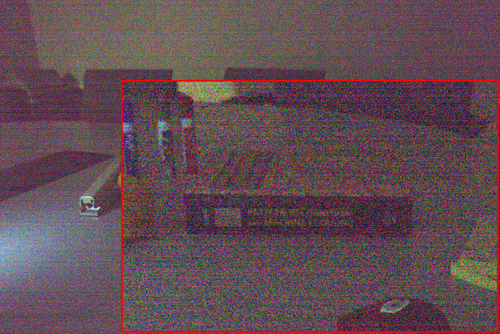} &
        \includegraphics[width=0.323\linewidth]{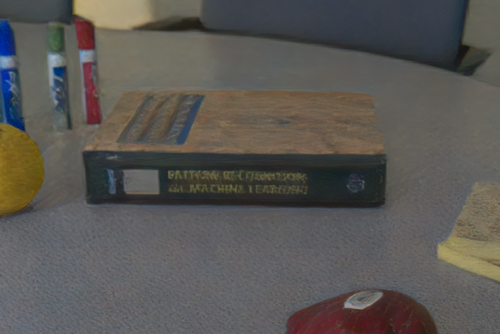} &
        \includegraphics[width=0.323\linewidth]{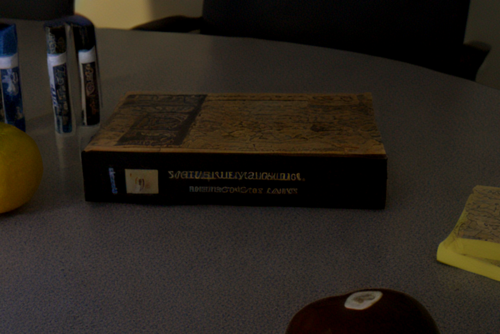} \\

        ELD~\cite{wei2020physics} & SID~\cite{chen2018learning} & LED~\cite{jiniccv23led} \\
        
        \includegraphics[width=0.323\linewidth]{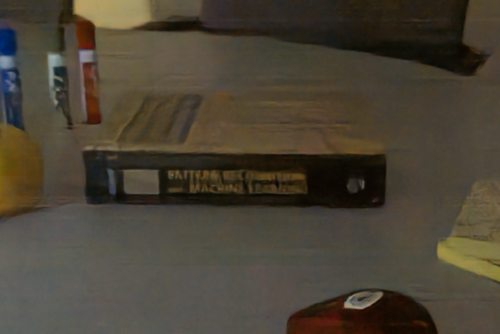} &
        \includegraphics[width=0.323\linewidth]{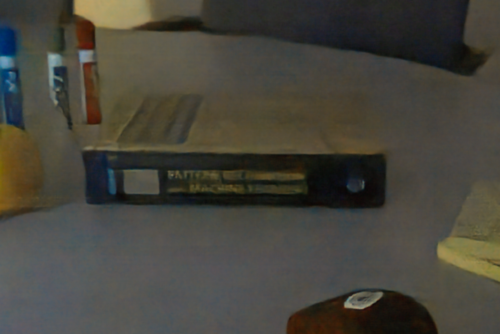} &
        \includegraphics[width=0.323\linewidth]{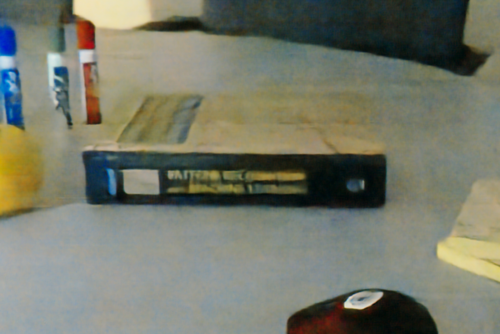} 
        \\

        ExposureDiffusion~\cite{wang2023exposurediffusion} & LRD~\cite{zhang2023towards} & Reference \\
        
        \includegraphics[width=0.323\linewidth]{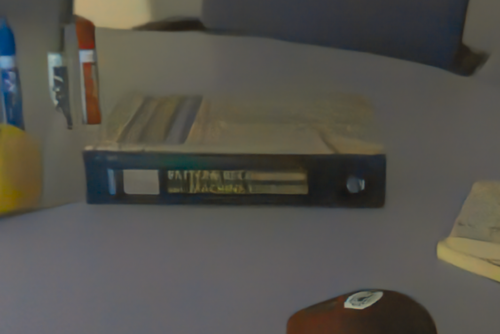} &
        \includegraphics[width=0.323\linewidth]{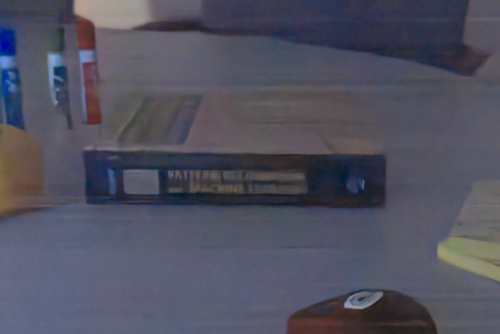} &
        \includegraphics[width=0.323\linewidth]{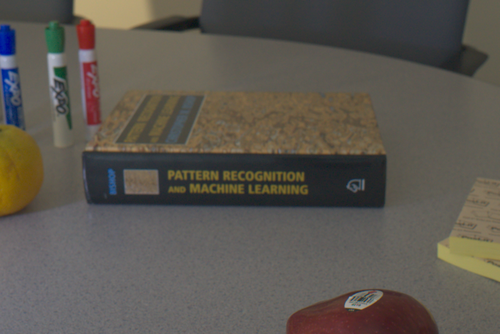} 
        \\

    \end{tabular}
    \caption{More qualitative results on SID dataset compared with baselines.
    }
    \label{fig:ablation_qual_1}
\end{figure*}

\begin{figure*}[t]
    \centering
    \begin{tabular}{c@{\hspace{1mm}}c@{\hspace{1mm}}c}

        Noisy Input & \ours & SD Concat \\

        \includegraphics[width=0.323\linewidth]{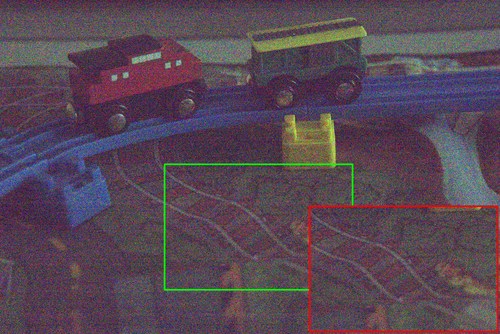} &
        \includegraphics[width=0.323\linewidth]{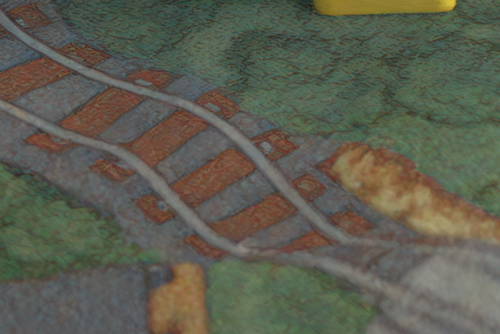} &
        \includegraphics[width=0.323\linewidth]{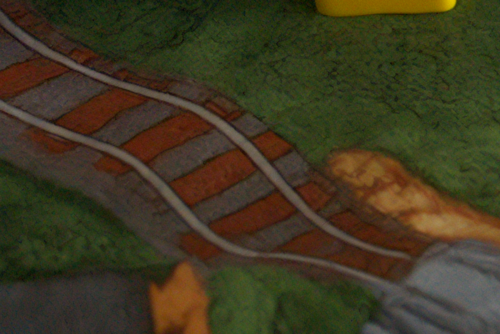} \\

        ELD~\cite{wei2020physics} & SID~\cite{chen2018learning} & LED~\cite{jiniccv23led} \\
        
        \includegraphics[width=0.323\linewidth]{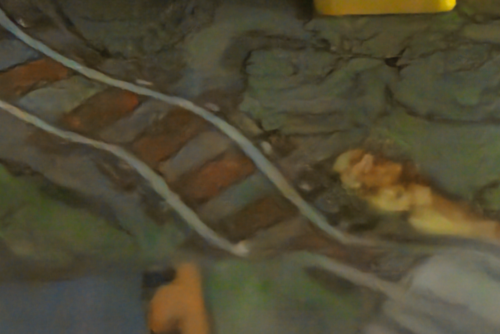} &
        \includegraphics[width=0.323\linewidth]{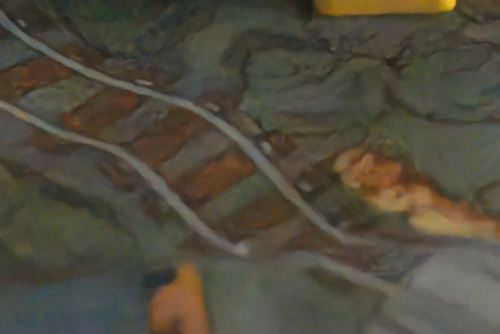} &
        \includegraphics[width=0.323\linewidth]{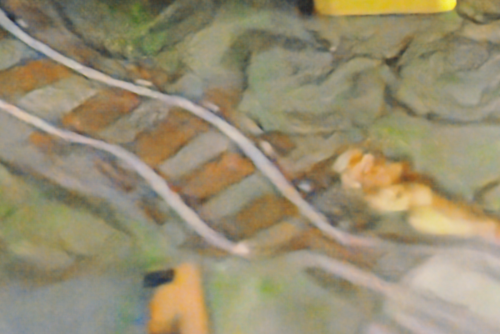} 
        \\

        ExposureDiffusion~\cite{wang2023exposurediffusion} & LRD~\cite{zhang2023towards} & Reference \\
        
        \includegraphics[width=0.323\linewidth]{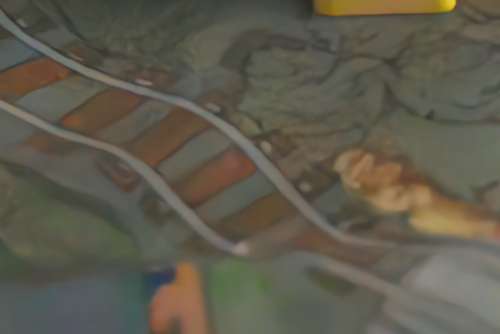} &
        \includegraphics[width=0.323\linewidth]{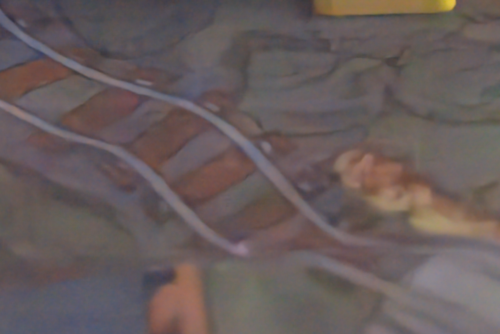} &
        \includegraphics[width=0.323\linewidth]{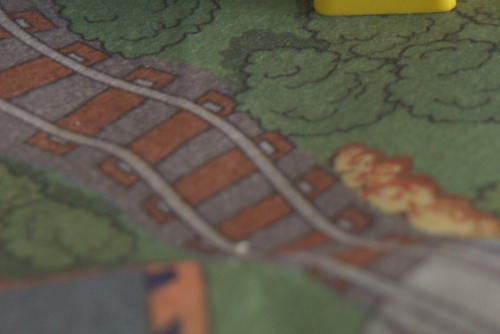} 
        \\

    \end{tabular}
    \caption{More qualitative results on the SID dataset compared with baselines.
    }
    \label{fig:ablation_qual_5}
\end{figure*}

\begin{figure*}[t]
    \centering
    \begin{tabular}{c@{\hspace{1mm}}c@{\hspace{1mm}}c}

        Noisy Input & \ours & SD Concat \\

        \includegraphics[width=0.323\linewidth]{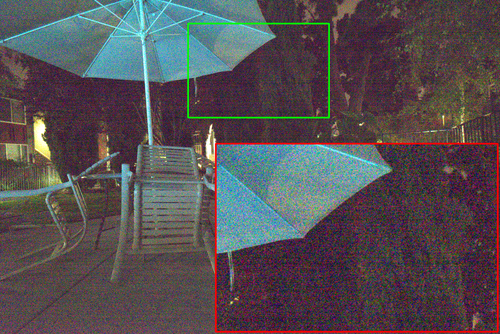} &
        \includegraphics[width=0.323\linewidth]{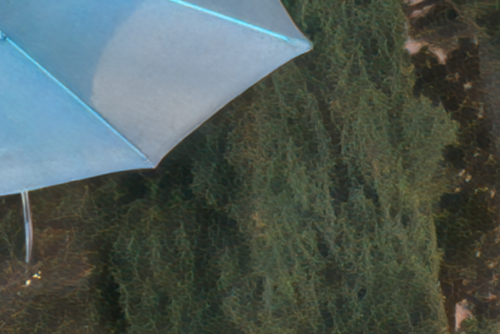} &
        \includegraphics[width=0.323\linewidth]{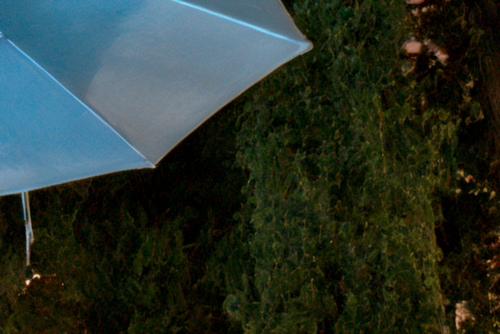} \\

        ELD~\cite{wei2020physics} & SID~\cite{chen2018learning} & LED~\cite{jiniccv23led} \\
        
        \includegraphics[width=0.323\linewidth]{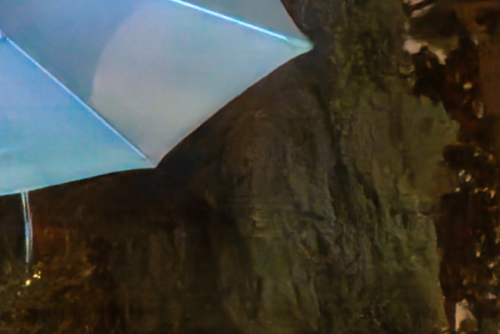} &
        \includegraphics[width=0.323\linewidth]{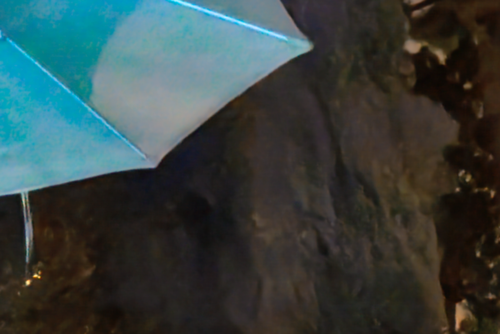} &
        \includegraphics[width=0.323\linewidth]{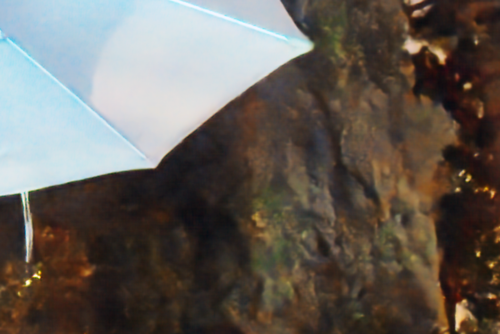} 
        \\

        ExposureDiffusion~\cite{wang2023exposurediffusion} & LRD~\cite{zhang2023towards} & Reference \\
        
        \includegraphics[width=0.323\linewidth]{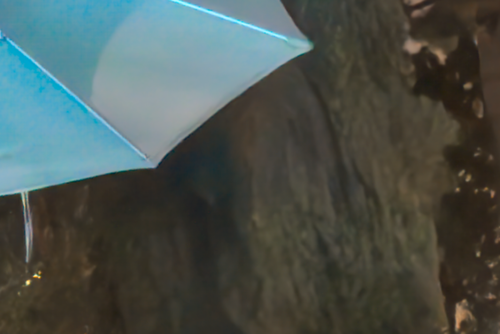} &
        \includegraphics[width=0.323\linewidth]{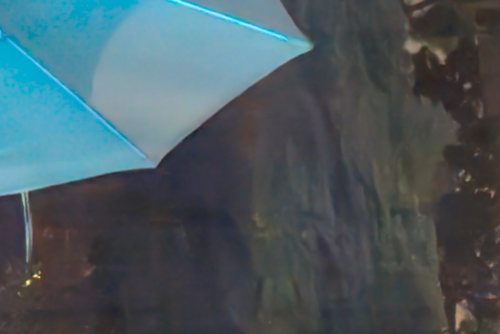} &
        \includegraphics[width=0.323\linewidth]{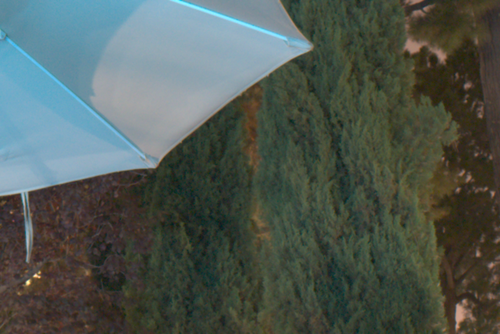} 
        \\

    \end{tabular}
    \caption{More qualitative results on the SID dataset compared with baselines.
    }
    \label{fig:ablation_qual_2}
\end{figure*}

\begin{figure*}[t]
    \centering
    \begin{tabular}{c@{\hspace{1mm}}c@{\hspace{1mm}}c}

        Noisy Input & \ours & SD Concat \\

        \includegraphics[width=0.323\linewidth]{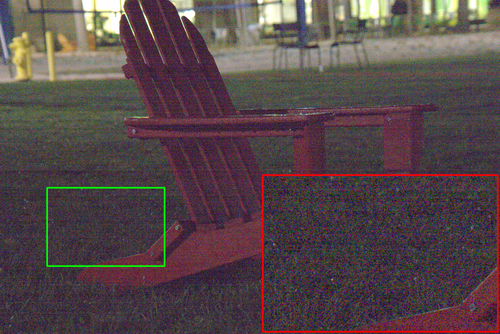} &
        \includegraphics[width=0.323\linewidth]{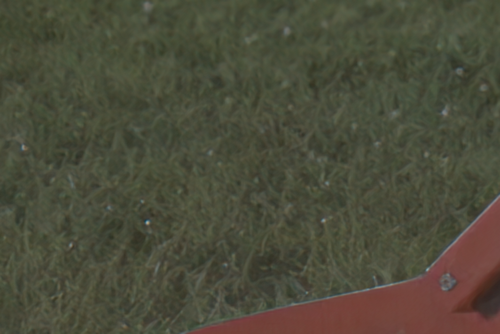} &
        \includegraphics[width=0.323\linewidth]{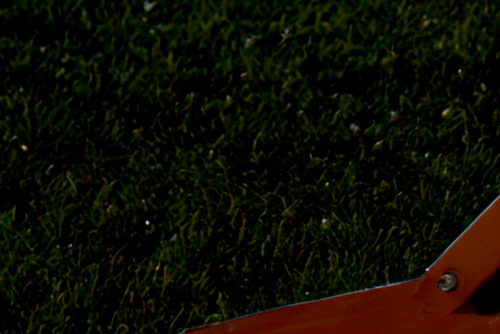} \\

        ELD~\cite{wei2020physics} & SID~\cite{chen2018learning} & LED~\cite{jiniccv23led} \\
        
        \includegraphics[width=0.323\linewidth]{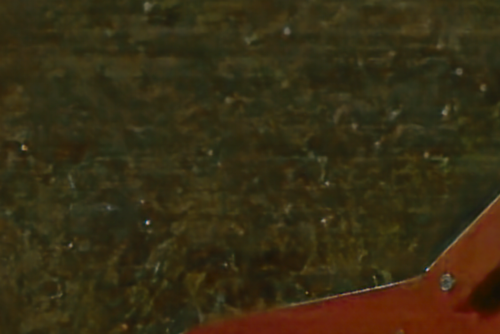} &
        \includegraphics[width=0.323\linewidth]{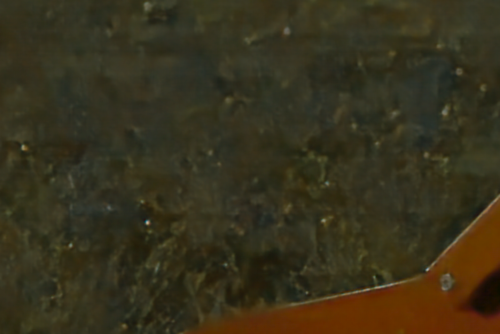} &
        \includegraphics[width=0.323\linewidth]{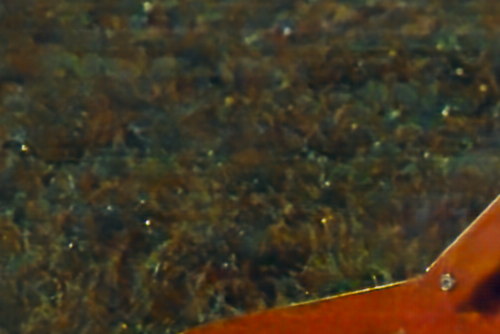} 
        \\

        ExposureDiffusion~\cite{wang2023exposurediffusion} & LRD~\cite{zhang2023towards} & Reference \\
        
        \includegraphics[width=0.323\linewidth]{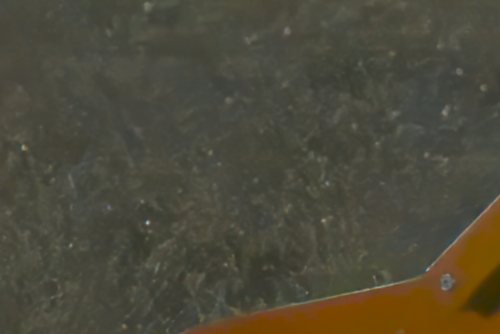} &
        \includegraphics[width=0.323\linewidth]{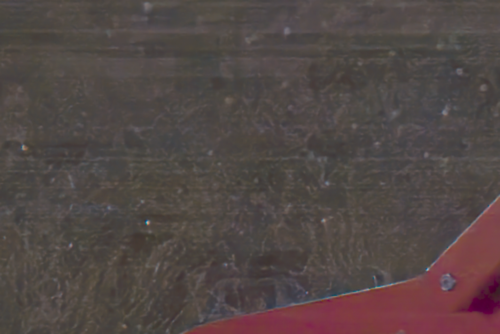} &
        \includegraphics[width=0.323\linewidth]{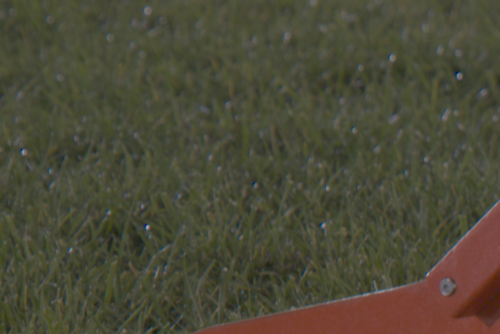} 
        \\

    \end{tabular}
    \caption{More qualitative results on the SID dataset compared with baselines.
    }
    \label{fig:ablation_qual_3}
\end{figure*}

\begin{figure*}[t]
    \centering
    \begin{tabular}{c@{\hspace{1mm}}c@{\hspace{1mm}}c}

        Noisy Input & \ours & SD Concat \\

        \includegraphics[width=0.323\linewidth]{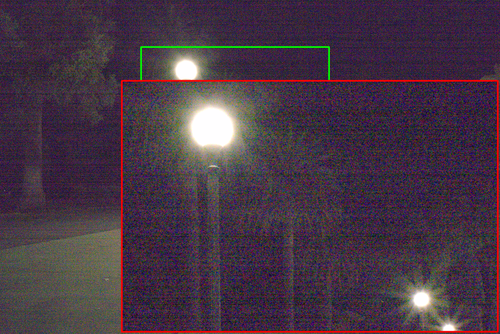} &
        \includegraphics[width=0.323\linewidth]{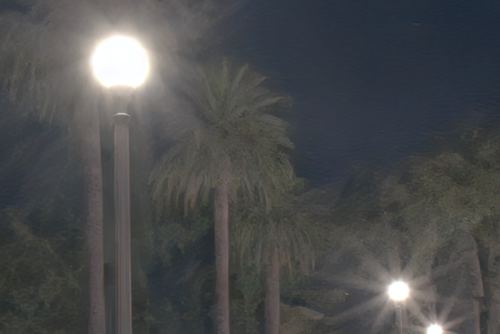} &
        \includegraphics[width=0.323\linewidth]{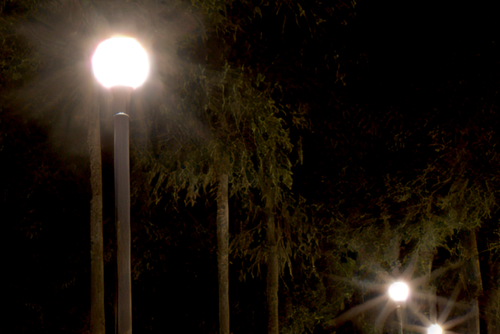} \\

        ELD~\cite{wei2020physics} & SID~\cite{chen2018learning} & LED~\cite{jiniccv23led} \\
        
        \includegraphics[width=0.323\linewidth]{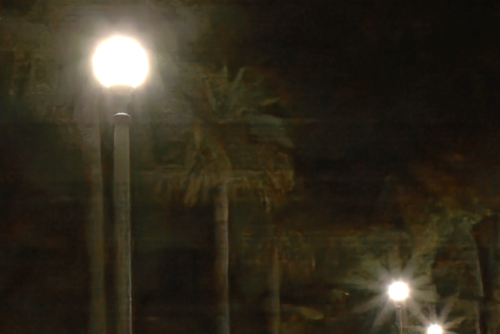} &
        \includegraphics[width=0.323\linewidth]{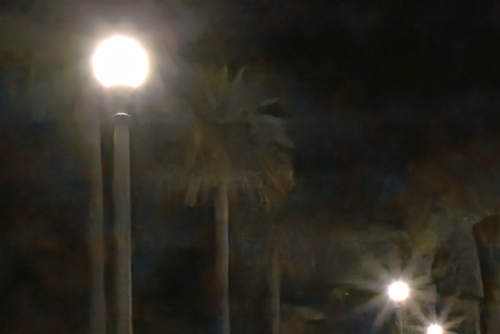} &
        \includegraphics[width=0.323\linewidth]{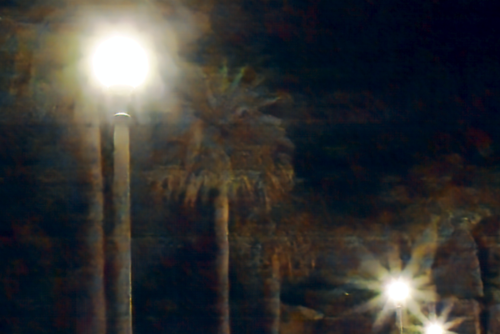} 
        \\

        ExposureDiffusion~\cite{wang2023exposurediffusion} & LRD~\cite{zhang2023towards} & Reference \\
        
        \includegraphics[width=0.323\linewidth]{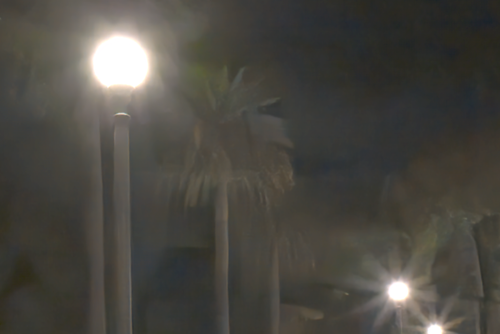} &
        \includegraphics[width=0.323\linewidth]{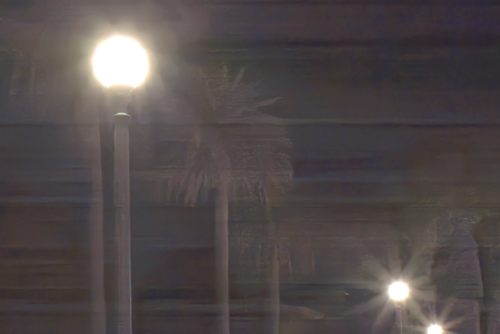} &
        \includegraphics[width=0.323\linewidth]{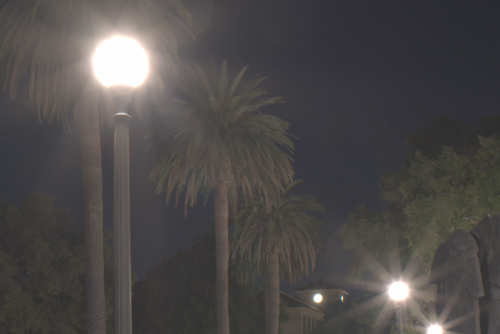} 
        \\

    \end{tabular}
    \caption{More qualitative results on the SID dataset compared with baselines.
    }
    \label{fig:ablation_qual_4}
\end{figure*}

We provide more visual results in this section. Specifically, we show the results compared with more baselines listed in~\tabref{tab:sid}. In~\figref{fig:ablation_qual_1} and~\figref{fig:ablation_qual_5}, we provide our indoor results and six baselines. As we can observe, only our results reconstruct the pattern of the book in~\figref{fig:ablation_qual_1}. The baselines either have color shifts or make the book pattern blurry. In~\figref{fig:ablation_qual_5}, we reconstruct the heavily noised rail while the baselines blur the rail. In~\figref{fig:ablation_qual_2},~\figref{fig:ablation_qual_3} and~\figref{fig:ablation_qual_4}, we show different outdoor scenes, including trees and grass. The baselines overly smoothed the images and lacked details. For example, in~\figref{fig:ablation_qual_2}, our result generates the leaves of the tree that resemble the reference image, while SD concat generates unrelated details, and the rest baselines simply smooth out the leaves. We provide the qualitative results in full images in the supplementary attachments.

\section{Implementation details}
\label{sec:implemtation_details}

\myparagraph{Hyperparameters.}
For stage 1, we train for $3,000$ epochs with a learning rate of $4e^{-5}$.
For stage 2, we train $3,000$ epochs with a learning rate of $2.5e^{-4}$ for the weights of the context processor and region-based cross-attention and $5e^{-5}$ for the rest of the weights. 
As we leverage classifier-free guidance in inference time, we randomly replace $5\%$ low-light noisy embeddings with pure Gaussian noise during stage-2 training. %
During training, we use image crops of size $1200\times 1200$ images. We run inference on the full image. 

\myparagraph{Model architecture.}
We use pre-trained Stable Diffusion V2-1 across all the experiments. Here we illustrate the architecture of the proposed path-wise cross-attention modules (\secref{sec:x-atten}) and residual VAE (\secref{sec:skip-vae}). For the region-based cross-attention layer, we keep the architecture and module sizes of the pre-trained Stable Diffusion unchanged and overwrite the \texttt{forward} function of the \texttt{SpatialTransformer} module to make it adapt to small patches instead of the whole input image embedding. %
For the Residual VAE, to match the shapes between the corresponding encoder and decoder features, we introduce Conv. layers in each block. The architecture details of the added layers are shown in~\tabref{tab:residual_vae_conv}.

\myparagraph{Data preprocessing pipeline.}
The raw input and target images are loaded and packed from the Bayer format into a 4-channel representation. The packing function extracts the RGB channels and normalizes the image using black and white levels. The pipeline for processing input raw Bayer images starts with a preprocessing step where white balance is applied to the raw RGBG data, followed by averaging the two green channels to produce the Linear RGB format. For the target images, the images are converted from raw to sRGB. The pipeline for converting raw Bayer images to sRGB images involves several key processing steps: white balance, binning, color correction, and gamma compression. First, white balance is applied by adjusting the pixel values based on pre-defined gains, ensuring accurate colors under varying lighting conditions. Then, the Bayer image undergoes binning, where the green channels are averaged and combined with the red and blue channels to form a linear RGB image. Color correction is performed by applying a color correction matrix that transforms the image into a more accurate color space. The image is then gamma-compressed to convert it from linear space to a gamma-corrected space, which adjusts the brightness and contrast for display devices.

\begin{table}[h]
    \setlength{\tabcolsep}{2pt}
\centering
\caption{Convolutional Layers in Residual VAE}
\begin{tabular}{ccccccc}
\hline
\textbf{Layer} & \textbf{Type} & \textbf{Kernel Size} & \textbf{Stride} & \textbf{Padding} & \textbf{In Channels} & \textbf{Out Channels} \\ \hline
1              & Conv2d        & 3 $\times$ 3         & 1               & 1                                  & 128                    & 256                       \\
2              & Conv2d        & 3 $\times$ 3         & 1               & 1                             & 256                    & 512                       \\ 
3            & Conv2d   & 3 $\times$ 3         & 1               & 1                              & 512                    & 512                       \\ 
4            & Conv2d   & 3 $\times$ 3         & 1               & 1                              & 512                    & 512                       \\ \hline
\end{tabular}
\label{tab:residual_vae_conv}
\vspace{-0.3cm}
\end{table}

\myparagraph{Stage 1 training.}
In stage 1, we fine-tune the residual VAE with the paired noisy lRGB and clean sRGB images. Specifically, we add residual connections between the embedding after each encoder block and the embedding before the corresponding decoder block, as illustrated in~\figref{fig:skip_connection_pipeline}. We then fine-tune the whole pre-trained VAE as well as the residual connections with Adam~\cite{kingma2014adam} for 3,000 epochs on both SID~\cite{chen2018learning} and LRD~\cite{zhang2023towards} where the learning rate is $4e^{-5}$ and the $\beta_1 = 0.5, \beta_2 = 0.9$. The training is conducted on 8 $\times$ A100 GPUS with 80GB memory.

\begin{figure*}[h]
    \centering
    \includegraphics[width=0.98\linewidth]{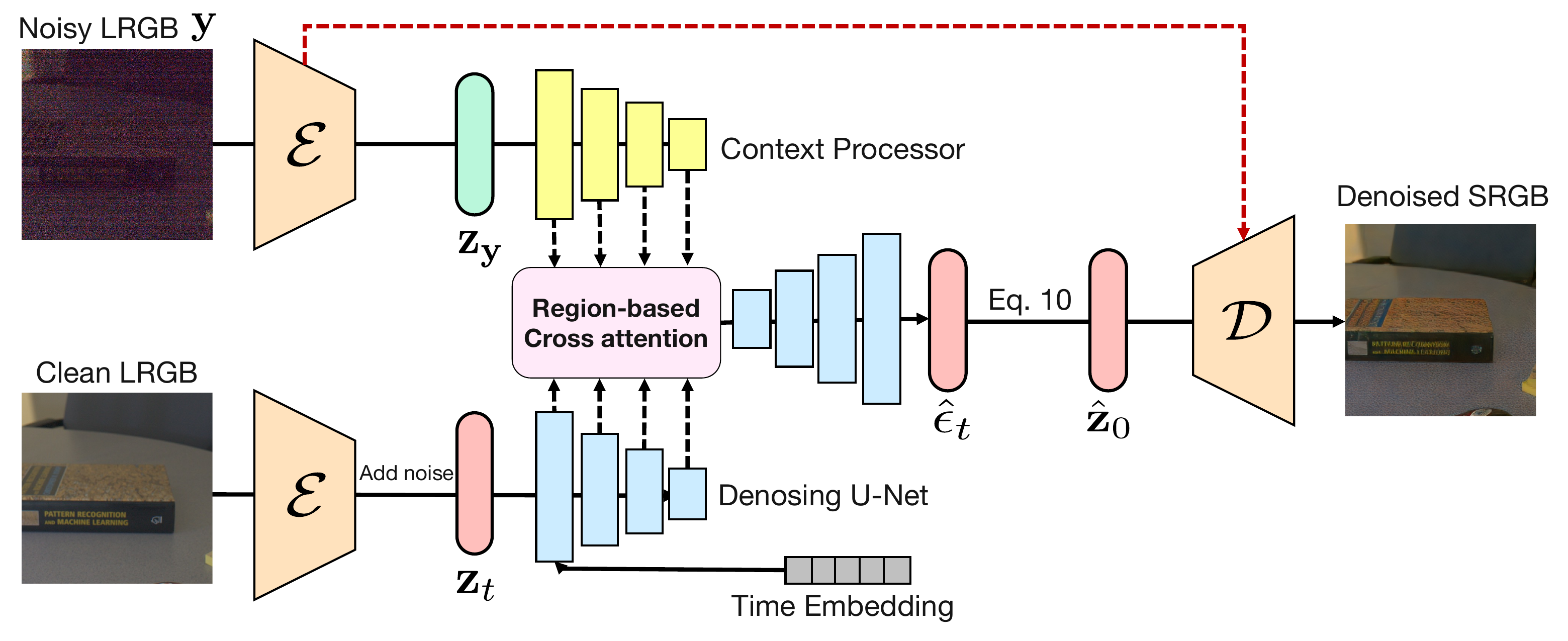}
    \caption{Training pipeline of Stage 2. We input noisy and clean linear RGB images into the same encoder to get $\rvz_\rvy$ and $\rvz_t$. The denoising U-Net outputs the estimated score $\hat \epsilon_t$, which is further transferred to $\hat \rvz_0$ based on~\equref{eq:latent_approx}, then the denoised output is used for computing the decoder-based reconstruction loss.}
    \label{fig:training_pipeline}
\end{figure*}

\myparagraph{Stage 2 training.}
In stage 2, we fine-tune the denoising network to incorporate the generative power of the pre-trained diffusion model. As shown in~\figref{fig:training_pipeline}, both the noisy and clean linear RGB images are input into the same trained VAE encoder from stage 1 to obtain $\rvz_\rvy$ and $\rvz_t$, where the context processor processes $\rvz_\rvy$ and $\rvz_t$ is processed by the U-Net encoder. The output of the denoising U-Net is transferred to $\hat \rvz_0$ and decoded by the trained VAE decoder from stage 1. The denoised sRGB output from the decoder is used to compute the decoder-based reconstruction loss defined in~\equref{eq:image_loss}. In our experiment, we replace the original cross-attention layers at each block of the original StableDiffusion, which are the layers across image and text embeddings, while we keep the cross-attention layers in the U-Net decoder and use the empty string as the text encoder input. Note that we use Xformers~\cite{xFormers2022} for all the attention modules. We fine-tune the whole U-Net, the context processor, and the region-based cross-attention layers with Adam~\cite{kingma2014adam} for 3,000 epochs. The training is conducted on 8 $\times$ A100 GPUS with 80GB memory.

\myparagraph{Inference.} 
At inference time, we use DDIM~\cite{song2020denoising} with 50 steps for sampling. As both SID\cite{chen2018learning} and ELD\cite{wei2020physics} datasets use a Sony A7S camera, we use the model trained on paired images from SID to do inference in ELD. The guidance weights is set to 2.0 for the SID~\cite{chen2018learning} and LRD datasets~\cite{chen2018learning}, and 2.5 for ELD datasets~\cite{wei2020physics}, respectively.

\myparagraph{Baselines.}
We illustrate the details of the baselines we introduced: SD Concat~\cite{brooks2023instructpix2pix} and SD Control~\cite{zhang2023adding,lin2023diffbir}. In SD Concat, we concat $\rvz_\rvy$ and $\rvz_t$ in the channel dimension and initialize corresponding weights for the additional channels in the first layer of the U-Net. In SD Control, we follow the structure of the ControlNet~\cite{zhang2023adding} that is a copy of the U-Net encoder and takes in the concatenation of $\rvz_\rvy$ and $\rvz_t$~\cite{lin2023diffbir}. The features from each layer are connected to the features of the corresponding U-Net decoder by using zero-conv layers.

\section{Limitations}
\label{sec:limitations}
The generation ability of the proposed method is constrained by that of the pre-trained diffusion model. We found that the current model has limited strength in enhancing non-English text in low-light images. In addition, diffusion models usually take much longer inference time than efficient convolutional networks. The proposed approach may need to run on cloud computing instead of on battery-limited devices for practical ISP processing. 

\end{document}